\documentclass{article}

\usepackage{amsmath,amsfonts,bm}

\def\eqref#1{equation~\ref{#1}}

\def\ceil#1{\lceil #1 \rceil}

\def\1{\bm{1}}

\def\vv{{\bm{v}}}

\def\mA{{\bm{A}}}

\def\mM{{\bm{M}}}

\DeclareMathAlphabet{\mathsfit}{\encodingdefault}{\sfdefault}{m}{sl}
\SetMathAlphabet{\mathsfit}{bold}{\encodingdefault}{\sfdefault}{bx}{n}
\newcommand{\tens}[1]{\bm{\mathsfit{#1}}}
\def\tA{{\tens{A}}}
\def\tB{{\tens{B}}}

\def\tG{{\tens{G}}}

\def\tQ{{\tens{Q}}}

\def\tX{{\tens{X}}}
\def\tY{{\tens{Y}}}

\def\sI{{\mathbb{I}}}

\newcommand{\R}{\mathbb{R}}

\usepackage{authblk}
\usepackage{arxiv}
\usepackage{subcaption}
\usepackage[table]{xcolor}
\usepackage[pagebackref=true,breaklinks=true,colorlinks,bookmarks=false]{hyperref}
\usepackage[capitalize]{cleveref}
\usepackage{amssymb}
\usepackage{natbib}
\usepackage{algorithm}
\usepackage{algcompatible}
\usepackage{mathtools}
\usepackage[binary-units=true]{siunitx}
\usepackage{multirow}
\usepackage{enumitem}
\usepackage{booktabs}

\newcommand{\fixedvspace}[1]{\par\kern-\prevdepth\vspace{#1}}

\setlist{nolistsep}

\newcommand{\acrotfourdt}{\textsf{T4DT}}
\title{\acrotfourdt{}: Tensorizing Time for Learning Temporal 3D Visual Data}

\author[1]{Mikhail Usvyatsov}
\author[2]{Rafael Ballester-Ripoll}
\author[3]{Lina Bashaeva}
\author[1]{Konrad Schindler}
\author[3]{Gonzalo Ferrer}
\author[3]{Ivan Oseledets}
\affil[1]{ETH~Zurich, Switzerland, \{mikhailu, schindler\}@ethz.ch}
\affil[2]{IE University,~Madrid, Spain, rafael.ballester@ie.edu}
\affil[3]{Skolkovo Institute of Science and Technology,~Moscow, Russia, \{lina.bashaeva, g.ferrer,  I.Oseledets\}@skoltech.ru }

\begin{document}
	
	\maketitle
	
	\begin{figure}[h!]
		\centering
		\begin{subfigure}[b]{0.24\textwidth}
			\includegraphics[width=\linewidth]{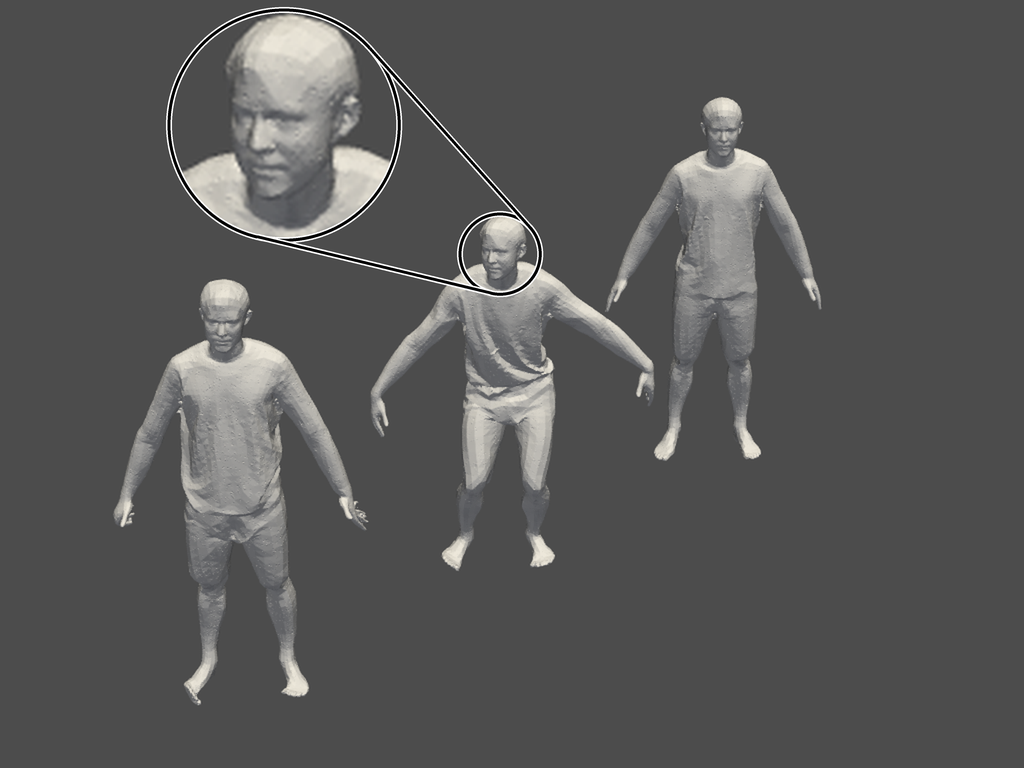}
			\caption{Original, \SI{284}{\giga\byte}}
			\label{fig:original_oqtt}
		\end{subfigure}
		\hfill
		\begin{subfigure}[b]{0.24\textwidth}
			\includegraphics[width=\linewidth]{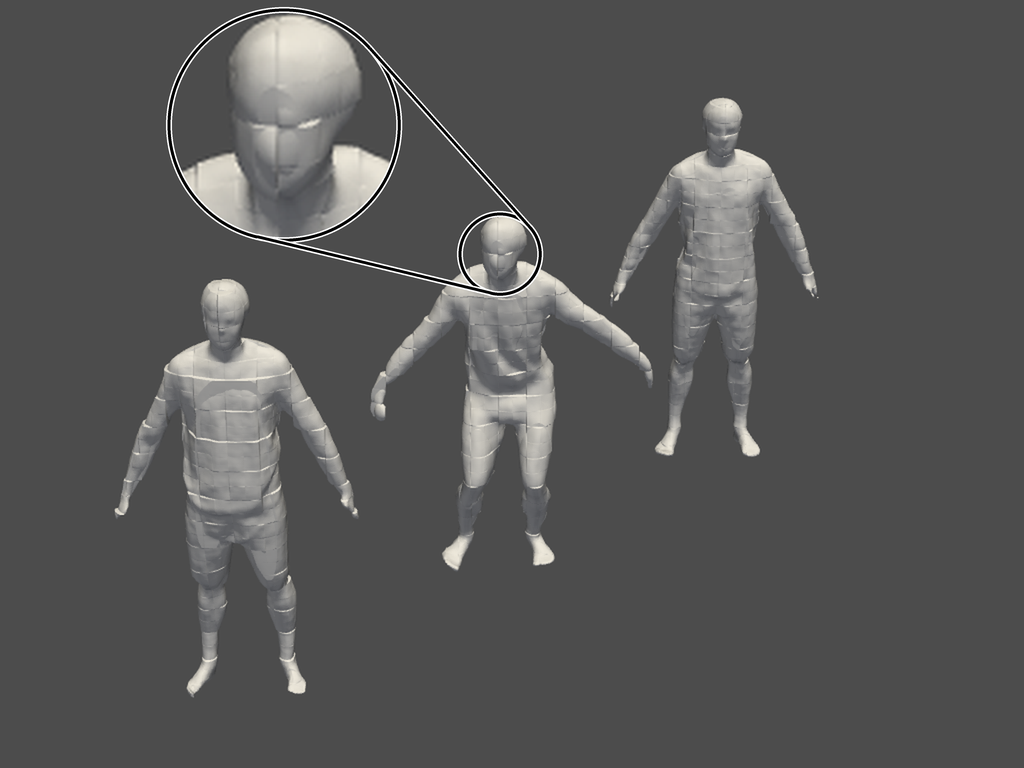}
			\caption{1:12345, \SI{24}{\mega\byte}}
			\label{fig:compressed_oqtt1}
		\end{subfigure}
		\hfill
		\begin{subfigure}[b]{0.24\textwidth}
			\includegraphics[width=\linewidth]{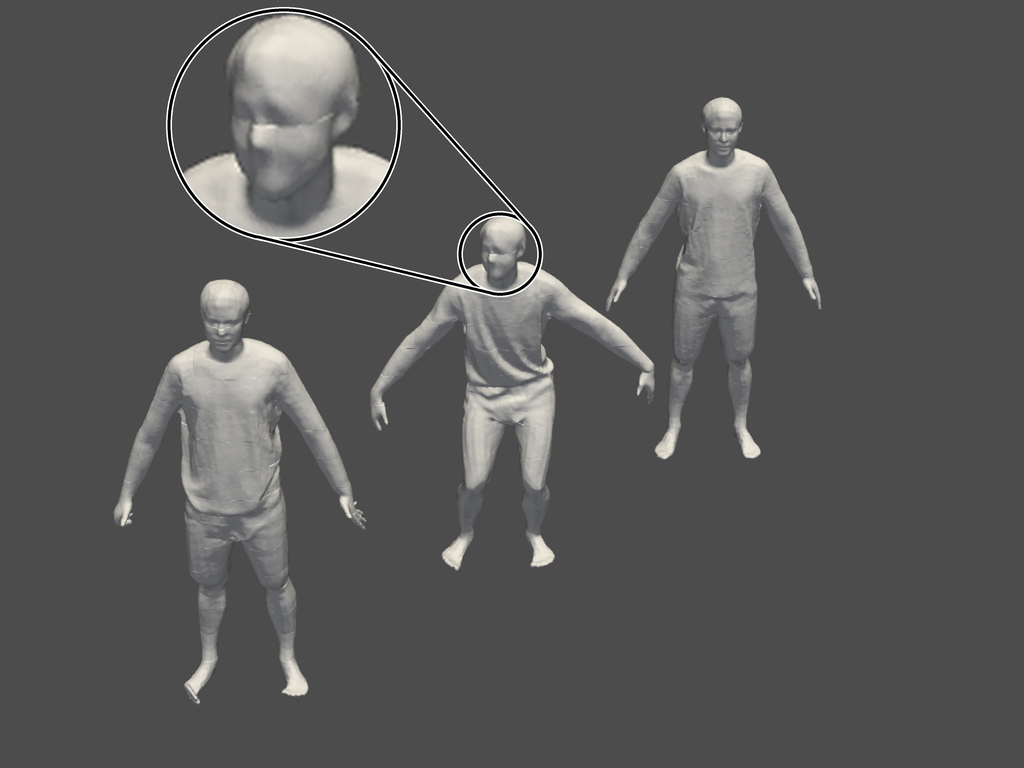}
			\caption{1:793, \SI{368}{\mega\byte}}
			\label{fig:compressed_oqtt2}
		\end{subfigure}
		\hfill
		\begin{subfigure}[b]{0.24\textwidth}
			\includegraphics[width=\linewidth]{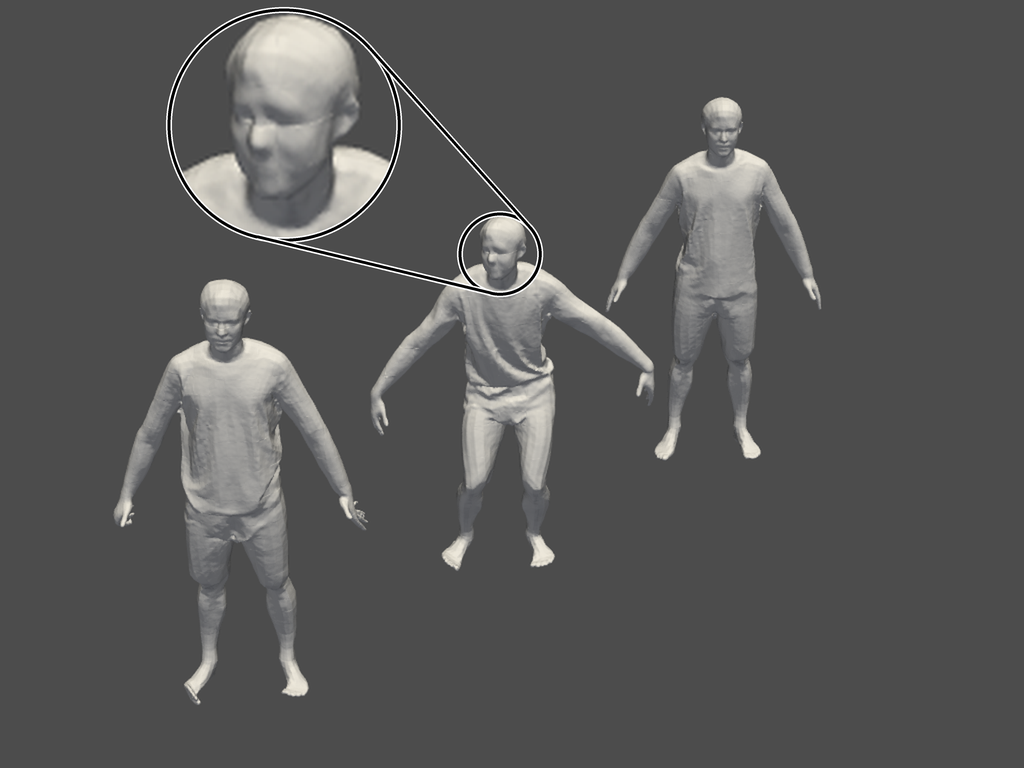}
			\caption{1:227, \SI{1.24}{\giga\byte}}
			\label{fig:compressed_oqtt3}
		\end{subfigure}
		\vspace{0.7em}
		\caption{\acrotfourdt{} with different compression levels in OQTT format for a \textit{longshort-flying-eagle} scene of resolution $512^3$ with 284 time frames. Only frames 1, 142, and 284 are depicted. The compression ratio is different from the actual memory consumption due to the padding of the time dimension to 512. High compression is achieved with $r_{\max} = 400$, $\text{MSDM2} = 0.45$ in \cref{fig:compressed_oqtt1}, medium compression with $r_{\max} = 1800$, $\text{MSDM2} = 0.32$ in \cref{fig:compressed_oqtt2}, and low compression / high quality with $r_{\max} = 4000$, $\text{MSDM2} = 0.29$ in \cref{fig:compressed_oqtt3}.}
		\label{fig:compression_oqtt}
	\end{figure}
	
	\begin{abstract}
		Unlike 2D raster images, there is no single dominant representation for 3D visual data processing. Different formats like point clouds, meshes, or implicit functions each have their strengths and weaknesses. Still, grid representations such as signed distance functions have attractive properties also in 3D. In particular, they offer constant-time random access and are eminently suitable for modern machine learning. Unfortunately, the storage size of a grid grows exponentially with its dimension. Hence they often exceed memory limits even at moderate resolution. This work proposes using low-rank tensor formats, including the Tucker, tensor train, and quantics tensor train decompositions, to compress time-varying 3D data. Our method iteratively computes, voxelizes, and compresses each frame's truncated signed distance function and applies tensor rank truncation to condense all frames into a single,  compressed tensor that represents the entire 4D scene.
		We show that low-rank tensor compression is extremely compact to store and query time-varying signed distance functions. It significantly reduces the memory footprint of 4D scenes while remarkably preserving their geometric quality. Unlike existing, iterative learning-based approaches like DeepSDF and NeRF, our method uses a closed-form algorithm with theoretical guarantees.
	\end{abstract}
	
	\section{Introduction}
	\label{sec:intro}

	Recent advances in hardware development have made it possible to collect rich depth datasets with commodity devices.
	For example, modern AR/VR hardware can record videos of 3D data, thus capturing their temporal evolution to yield 4D datasets.
	Nevertheless, working with 4D data presents computational challenges due to the curse of dimensionality.
	Furthermore, while 2D data mostly come in the form of raster images, there is an entire zoo of popular representations for 3D data, ranging from point clouds and meshes to voxel grids and implicit (neural) functions.
	
	In this work, we address the case of representations based on regular grids. Such representations are highly structured and allow for efficient manipulation, e.g., they offer random access, slicing, and other operations in constant time. However, their storage requirements become a bottleneck: a grid of resolution $I$ along each dimension $D$ has $I^D$ elements.
	We propose applying tensor decompositions to leverage the high temporal and spatial correlation in grids that arise from time-evolving 3D data.
	Besides exploring the general idea of compression via low-rank constraints, we compare different tensor decomposition schemes and their potential for 4D video.
	For example, although the Tucker format~\cite{tucker1963implications,tucker1966some} is known to yield good results for the 3D case, it still suffers from the curse of dimensionality since it requires $O(r^D)$ elements w.r.t. the Tucker rank $r$, while $r$ must typically be chosen $\approx\frac{I}{\text{const.}}$ to achieve reasonable accuracy.
	Therefore, we explore hybrid low-rank formats based on the tensor train (TT) decomposition~\cite{oseledets2011tensor}.

	The Signed Distance Function (SDF, sometimes also called Signed Distance Field) and its truncated version, known as Truncated SDF or TSDF, is a convenient format for data fusion: several methods convert partial 3D views, even if initially in mesh format, to TSDFs to fuse them into complete models \cite{Izadi2011KinectFusionR3}. This pipeline is useful for multi-view stereo acquisition, laser scanners, as well as resource-limited devices (e.g., mobile augmented reality).
	%
	TSDFs offer 1-to-1 correspondence across time, which is not easy to achieve with meshes if the objects undergo non-rigid deformation; while every mesh can be converted to a TSDF on the same voxel grid (with appropriately chosen resolution to avoid loss of detail).
	We experimentally analyze the effects of the low-rank constraint, using TSDF sequences of 3D scenes from the CAPE dataset \cite{CAPE:CVPR:20,pons2017clothcap} and problem-specific quality metrics. Our method can compress time-varying grids to a usable size, while the same scenes in uncompressed form would each require hundreds of GigaBytes of memory.
	
	In the following, \cref{sec:related} reviews applicable tensor methods and techniques for 3D and 4D data compression. \cref{sec:method} gives a detailed outline of our proposed approach, \cref{sec:results} demonstrates its the performance in an experimental study, followed by a discussion of limitations in \cref{sec:limitations}, and concluding remarks in \cref{sec:conclusions}.
	Our contributions are:
	\begin{enumerate}
		\item The first tensor decomposition framework for compressing temporal sequences in 3D voxel space;
		\item A benchmark and analysis of different decomposition schemes for both the spatial and temporal dimensions;
		\item An open source implementation of our method based on the \textsf{tntorch} framework \cite{UBS:22}, available at \url{https://github.com/prs-eth/T4DT}.
	\end{enumerate}
	
	\section{Background and Related Work}
	\label{sec:related}
	In our work, we apply tensor methods to temporal sequences of 3D scenes.
	We consider grid representations of the 3D data, i.e., a tensor $\displaystyle \tA \in \R^{I_1 \times \dots \times I_D}$ constitutes a discrete sampling of a $D$-dimensional space on a grid $\mathbb{I} = I_1 \times \dots \times I_D$, with $I_d$ samples along dimension $d$.
	
	\subsection{SDF and Truncated SDF}
	\label{subsec:tsdf}
	
	The SDF at position $p\in \R^3$ is defined as:
	\begin{equation}
		\label{eq:sdf}
		\text{SDF}(p) =
		\begin{cases}
			\text{dist}(p, \partial \Omega) & \text{if } p \in \Omega, \\
			\hfill -\text{dist}(p, \partial \Omega) & \text{otherwise},
		\end{cases}
	\end{equation}
	while the truncated SDF clamps the SDF as follows:
	\begin{equation}
		\label{eq:tsdf}
		\text{TSDF}(p) =
		\begin{cases}
			\hfill -\tau & \text{if } \text{SDF}(p) \leq -\tau, \\
			\hfill \tau & \text{if } \text{SDF}(p) \geq \tau, \\
			\hfill \text{SDF}(p) & \text{otherwise}.
		\end{cases}
	\end{equation}
	
	Depending on the available input data and the desired application, different formats may be more or less suitable. For example, colored images can be converted to NERF or NELF \cite{mildenhall2020nerf, sitzmann2021light}, which optimizes the reconstruction error via differentiable rendering. Depth images can be fused into TSDFs as described in \cite{curless1996volumetric}; meshes can be converted into TSDFs, too, and also the inverse transformation is possible with variants of marching cubes~\cite{lorensen1987marching}; point clouds can be turned into meshes or directly into TSDFs; etc.
	
	\subsection{Compression}
	\label{subsec:compression}
	Storing signed distance fields constitutes a computational and memory bottleneck, especially for time-varying data.
	In \cite{tang2018real}, a real-time compression method was proposed that uses a learnable, implicit temporal TSDF representation, combining independent color and texture encodings with learnable geometry compression.
	Multi-dimensional visual data compression based on the Tucker decomposition, combined with bit-plane coding, was applied to volumetric data in \cite{ballester2019tthresh}.
	Later,~\cite{boyko2020tt} proposed compressing single-frame TSDFs with a block-based neural network. They showed that, although TSDF tensors usually cannot be written exactly as a low-rank expansion, in practice, one can store them in the low-rank TT representation with low compression error.
	Here, we build on the idea of tensor decompositions as a tool for compressing high-dimensional visual data. We address the task of compressing dynamic 3D scenes, whose 4D grid representations quickly exceed practical memory limits when stored in uncompressed form.
	
	Compressing TSDF with tensor methods not only allows one to work with data that otherwise would not fit the memory, but it can also be useful for:
	\begin{enumerate}
		\item Fusion, as demonstrated in \cite{boyko2020tt};
		\item Constant-time TSDF random access,\footnote{The complexity of accessing a single element depends on the choice of tensor format and rank but is independent of the location of the query point within the tensor} which may be useful to test if a given point is inside or outside of a mesh, and potentially for ray-casting;
		\item Constant-time computation of the TSDF gradient, used to compute surface normals~\cite{sommer2022gradient};
		\item Unlike neural network-based learning approaches, tensor methods rely on SVD-based schemes, which provide theoretical guarantees~\cite{oseledets2011tensor} and make the framework more interpretable.
	\end{enumerate} 
	
	Throughout this work, we use the term \textit{compression ratio} (or just \textit{compression}, if clear from the context) to denote the ratio between the number of coefficients used for the compressed format and the number of coefficients required to store the uncompressed tensor.
	
	\section{Tensor Formats}
	\label{sec:tensor_formats}
	
	\subsection{Notation}
	\label{subsec:notation}
\begin{figure}[h!]
		\centering
		\begin{subfigure}[b]{0.3\textwidth}
			\hspace{3.5em}
			\includegraphics[scale=1.]{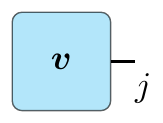}
			\caption{vector}
			\label{fig:vector}
		\end{subfigure}
		\hfill
		\begin{subfigure}[b]{0.3\textwidth}
			\hspace{2.5em}
			\includegraphics[scale=1.]{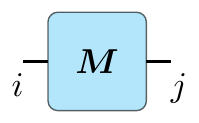}
			\caption{matrix}
			\label{fig:matrix}
		\end{subfigure}
		\hfill
		\begin{subfigure}[b]{0.3\textwidth}
			\hspace{0.5em}
			\includegraphics[scale=1.]{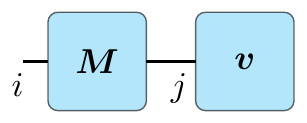}
			\caption{matrix-vector product}
			\label{fig:matrix_vector}
		\end{subfigure}
		\vspace{0.5em}
		\caption{Each tensor is denoted as a node whose edges represent its dimensions. Whenever an edge is shared, tensor contraction is assumed: e.g., \cref{fig:matrix_vector} depicts the matrix-vector product $\sum_j \mM_{ij} \vv_j$.}
		\label{fig:diagram_notation}
	\end{figure}

	\paragraph{Tensor diagram notation.}
	To visualize tensor networks, we adopt Penrose's graphical notation \cite{penrose1971applications}. See \cref{fig:diagram_notation} for visualization of a vector, a matrix and a matrix-vector product.
	
	\subsection{Tucker}
	\label{subsec:tucker}
	
\begin{figure}[h!]
		\centering
		\begin{subfigure}[b]{\textwidth}
			\centering
			\includegraphics[scale=1.]{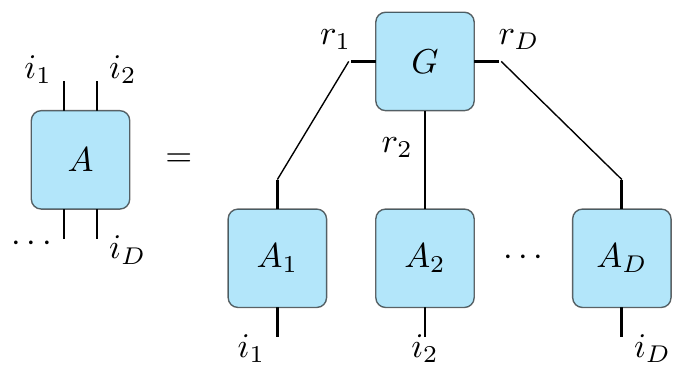}
			\caption{Tucker}
			\label{fig:tucker}
		\end{subfigure}
		\begin{subfigure}[b]{\textwidth}
			\centering
			\includegraphics[scale=1.]{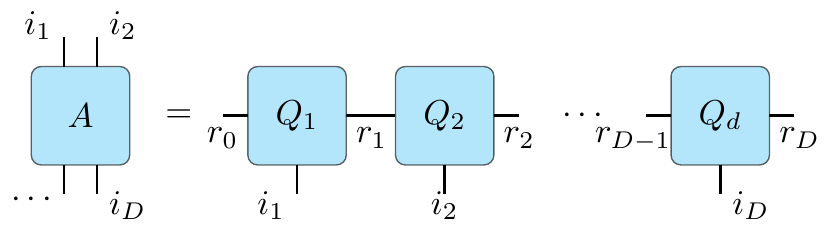}
			\caption{TT}
			\label{fig:tt}
		\end{subfigure}
		\caption{Graphical representation of the Tucker~\cref{fig:tucker} and TT~\cref{fig:tt} decompositions.}
		\label{fig:decompositions}
	\end{figure}

	The Tucker decomposition~\cite{tucker1963implications} factors a tensor of dimension $D$ into a $D$-dimensional \emph{core} tensor $\displaystyle \tG \in \R^{r_1 \times \dots \times r_D}$ and $D$ matrices $\displaystyle \{\mA_d\}_{d=1}^{D}$, $\displaystyle \mA_d \in \R^{r_d \times I_d}$.
	The format is defined as
	\begin{equation}
		\label{eq:tucker}
		\displaystyle
		\tA[i_1, \dots, i_D] = 
		\tG
		\tA_1[:, i_1]
		\dots
		\tA_D[:, i_D],
	\end{equation}
	where $r_d$ are the Tucker-ranks.
	The Tucker decomposition has $\mathcal{O}\big((\max_d[r_d])^D + \max_d[r_d]\cdot\max_d[I_d]\big)$ storage cost. See \cref{fig:tucker} for visualization of Tucker format in graphical notation.
	
	\subsection{Tensor Train}
	\label{subsec:tt}
	
	The TT decomposition~\cite{oseledets2011tensor} factors a tensor of dimension $D$ into a sequence of $D$ 3-dimensional tensors.
	The format is defined as
	\begin{equation}
		\label{eq:tt}
		\displaystyle
		\tA[i_1, \dots, i_D] = 
		\tQ_1[0, i_1, :]
		\tQ_2[:, i_2, :]
		\dots
		\tQ_D[:, i_D, 0],
	\end{equation}
	where the tensors $\displaystyle \{\tQ_d\}_{d=1}^D, 
	\tQ_d \in \R^{r_{d-1} \times I_d \times r_d}$, are called \textit{TT-cores}; and $r_d$ are the \emph{TT-ranks} ($r_0\!=\!r_D\!=\!1$).
	The TT decomposition has $\mathcal{O}\big(D\cdot(\max_d[r_d])^2\cdot\max_d[I_d]\big)$ storage cost, and leads to a linear tensor network; see \cref{fig:tt} for a graphical example.
	
	\subsection{QTT}
	\label{subsec:qtt}
	Quantics TT (QTT) is an extension of TT that includes reshaping\footnote{With zero-padding where needed} the input into a tensor of shape $2\times2 \dots \times 2 = 2^{\sum_{d=1}^{D} \log_2(I_d)}$, with $I_j$ the size of the $j$-th mode \cite{kazeev2016qtt}.
	The sub-dimensions $x, y, z$ are then grouped side-by-side for each octet, which imitates the traversal of a $z$-space filling curve and makes QTT similar to a tensorized octree. Last, the resulting tensor is then subject to standard TT decomposition. This scheme is also connected to the wavelet transform~\cite{kazeev2013tensor}.
	
	\paragraph*{Octet QTT.} We also propose to reuse a variant of QTT with a base dimension of size eight instead of two by merging the three sub-dimensions of each octet into one. This format was originally proposed by \citet{qttnf2022}. In this way, the resulting tensor has shape $8 \times 2 \times ... \times 8 \times 2$.
	We refer to this format as OQTT for octet QTT and found it to reduce discontinuity artifacts (\cref{sec:results}).
	
	\section{Proposed Method}
	\label{sec:method}
	
	We introduce \acrotfourdt{}, a method to compress high-resolution temporal TSDF fields with tensor decompositions discussed in Sections \ref{subsec:tt} - \ref{subsec:qtt}.
	We exploit that individual TSDF frames can be compressed into a low-rank TT decomposition with reasonable error~\cite{boyko2020tt}. Under the low-rank constraint, a zero-level set can be reconstructed with sufficiently good quality.
	However, this becomes more challenging when considering time-evolving data since uncompressed 4D grid scenes at fine spatial resolutions can range in the hundreds of GBs. \cref{alg:t4dt_pipeline} gives a high-level overview of our pipeline.
	
	\begin{algorithm}[h!]
		\label{alg:t4dt_pipeline}
		\begin{algorithmic}
	\STATEx $\sI = I_1 \times I_2 \times I_3 \times I_4$ -- temporal 3D grid 
	\STATEx $\tX \in \R^{\sI}$ -- input temporal TSDF
	\STATEx $R_s$ -- maximal rank along spatial dimensions
    \STATEx $R_t$ -- maximal rank along time dimension
    \STATEx $\displaystyle \tX^\prime$ -- Collection of compressed TSDF frames
    \STATEx $\displaystyle \tY$ -- Compressed temporal TSDF
	\REQUIRE

	\FOR{$\text{i} = 1, 2, \cdots, I_4 - 1,  I_4$}
	\STATE $\displaystyle \tX^\prime_{\text{i}} \leftarrow$ truncate($\displaystyle \tX[\cdots, \text{i}], R_s$)
	\ENDFOR
	
	\WHILE{$\displaystyle \tX^\prime \neq \varnothing$ \textbf{and} $|\tY| \neq 1$}
	\STATE $\displaystyle \tY \leftarrow \text{truncate}(\tX_0^\prime \cup \tY, R_t)$
	\STATE $\text{remove}(\tX_0^\prime)$
	\ENDWHILE
\end{algorithmic}

		\caption{\acrotfourdt{} pipeline. Since the whole 4D scene is not expected to fit into memory, we first compress each frame individually. Next, the collection of individually compressed frames is assembled into a single compressed scene using a tree-like merge procedure $\tX_0^\prime \cup \tY$. See \cref{subsec:frame_concatenation} for the merge procedure details.}
	\end{algorithm}
	
	\begin{figure}[h!]
		\centering
		\includegraphics[width=\linewidth]{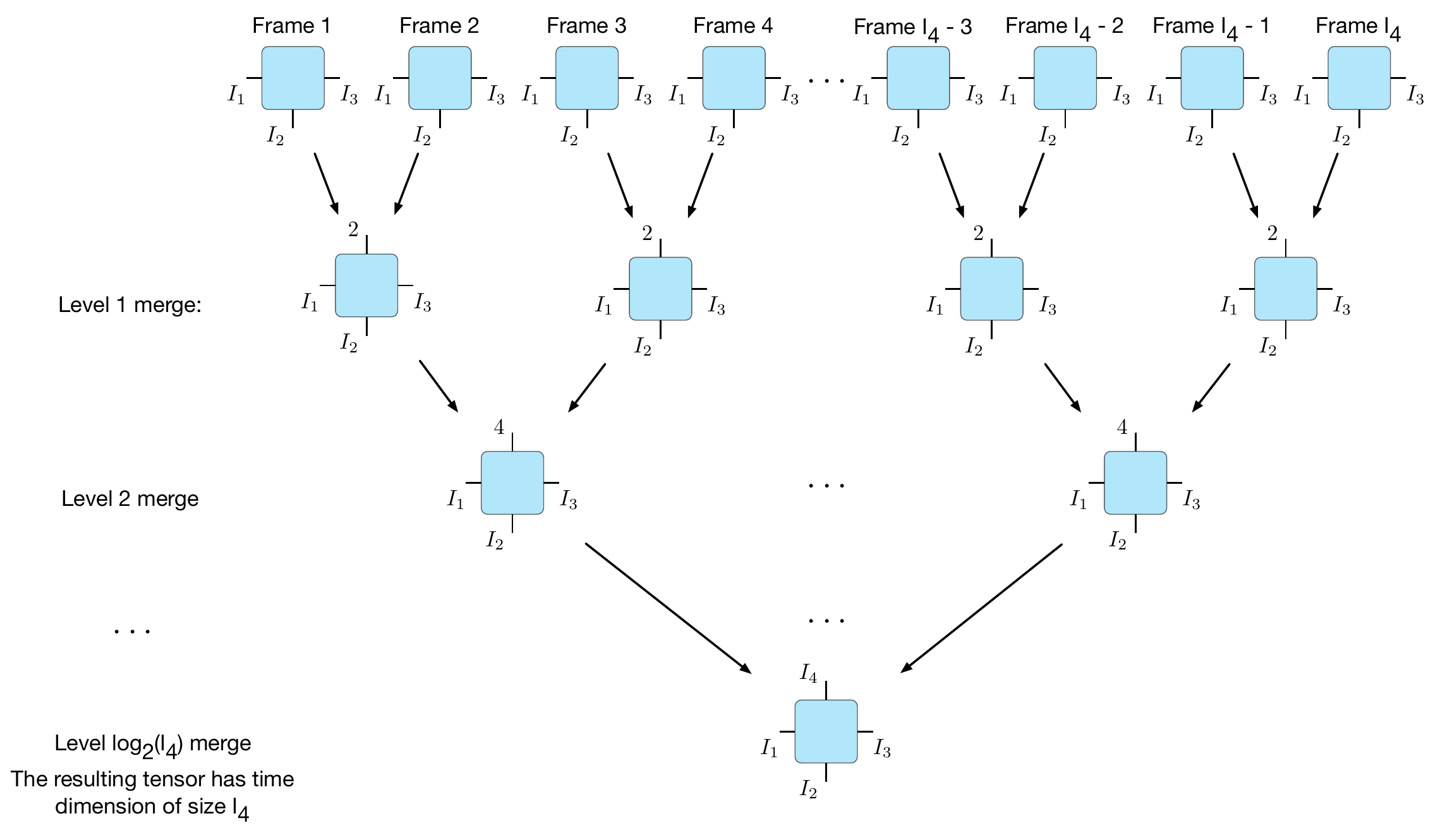}
		\caption{We compress and merge individual frames progressively into a compressed scene (for simplicity, all tensors are shown in uncompressed form). Two of the previous level tensors are merged at each tree level using the concatenation procedure (which increases tensor rank), followed by rank truncation. On the first level, an additional core for the temporal dimension is inserted into each frame.}
		\label{fig:frame_concatenation}
	\end{figure}
	
	\subsection{Framewise Decomposition}
	The first technical choice is the base decomposition at the frame level. We considered three variants: Tucker, TT, and QTT.
	In 3D, Tucker is an attractive choice since it is equivalent to TT plus an additional compression step along the second dimension. For the 4D case, we propose to combine the best of both worlds by using a TT-Tucker blend, where spatial dimensions are stored in Tucker formatted cores, and the temporal dimension is represented with a TT core.
	Each core or factor is responsible for a single dimension in both the TT and Tucker formats. However, the Tucker model shares the core tensor between all dimensions.
	
	Many real-world datasets are sparse in the sense that most of the occupied volume is empty.
	An octree is an excellent choice to pack such sparse data into a compact representation.
	We argue that QTT is a tensor analog of an octree data structure.
	Indeed, it is easy to see that, padded and reshaped to $2^{\sum_{d=1}^{D} \log_2(I_d))}$, 
	the scene becomes a piece-wise separated, reshaped set of octets after permutation of the corresponding sub-dimensions. Rank truncation is a way to reduce the number of coefficients to encode each octet sub-dimension.
	
	\subsection{Frame Concatenation}
	\label{subsec:frame_concatenation}
	Since the original scene can by far exceed the available memory, we developed a progressive procedure to merge compressed frames into a single compressed scene. The algorithm is akin to the \emph{pairwise summation} method to reduce round-off errors in numerical summation~\cite{Higham:02}: we stack the frames by pairs along the time dimension, recompress each pair via tensor rank truncation~\cite{oseledets2011tensor}, and repeat the procedure recursively in a binary-tree fashion to ultimately yield a single 4D compressed tensor. See \cref{fig:frame_concatenation} for an illustration.
	
	\begin{figure}[h!]
		\centering
		\begin{subfigure}[b]{0.49\textwidth}
			\includegraphics[width=\linewidth]{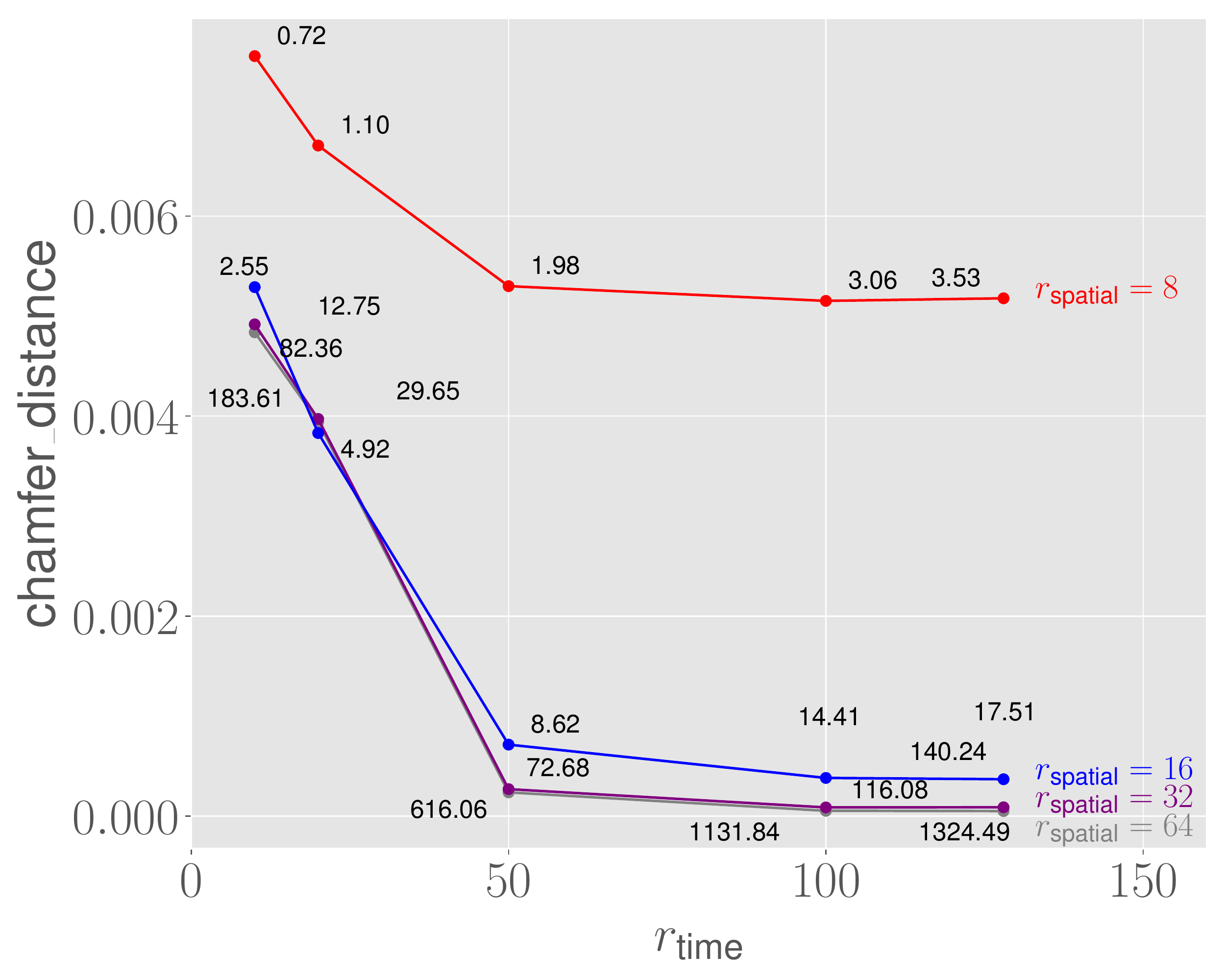}
			\caption{Chamfer distance}
			\label{fig:metric4}
		\end{subfigure}
		\begin{subfigure}[b]{0.49\textwidth}
			\includegraphics[width=\linewidth]{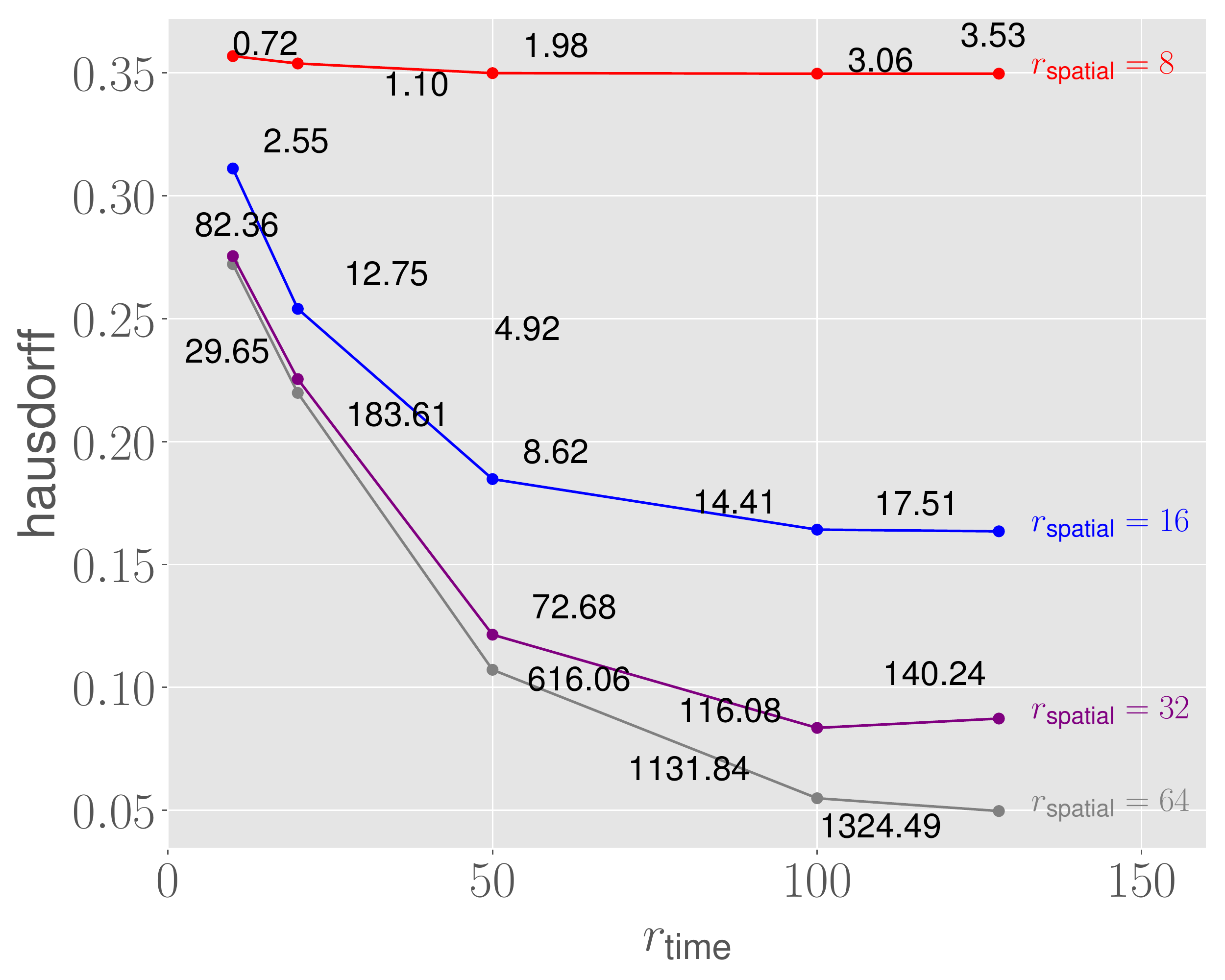}
			\caption{Hausdorff distance}
			\label{fig:metric3}
		\end{subfigure}
		\begin{subfigure}[b]{0.49\textwidth}
			\includegraphics[width=\linewidth]{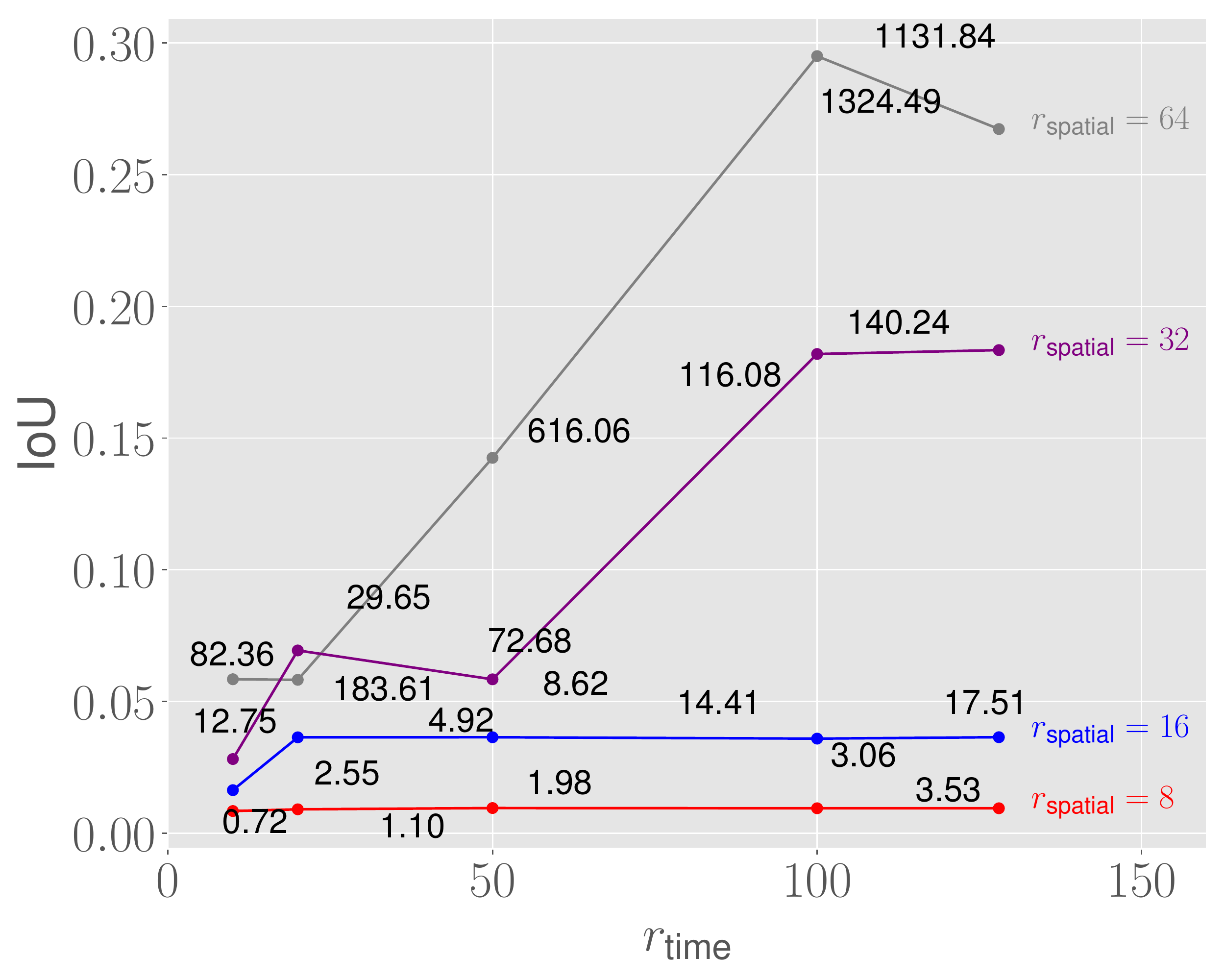}
			\caption{IoU}
			\label{fig:metric2}
		\end{subfigure}
		\begin{subfigure}[b]{0.49\textwidth}
			\includegraphics[width=\linewidth]{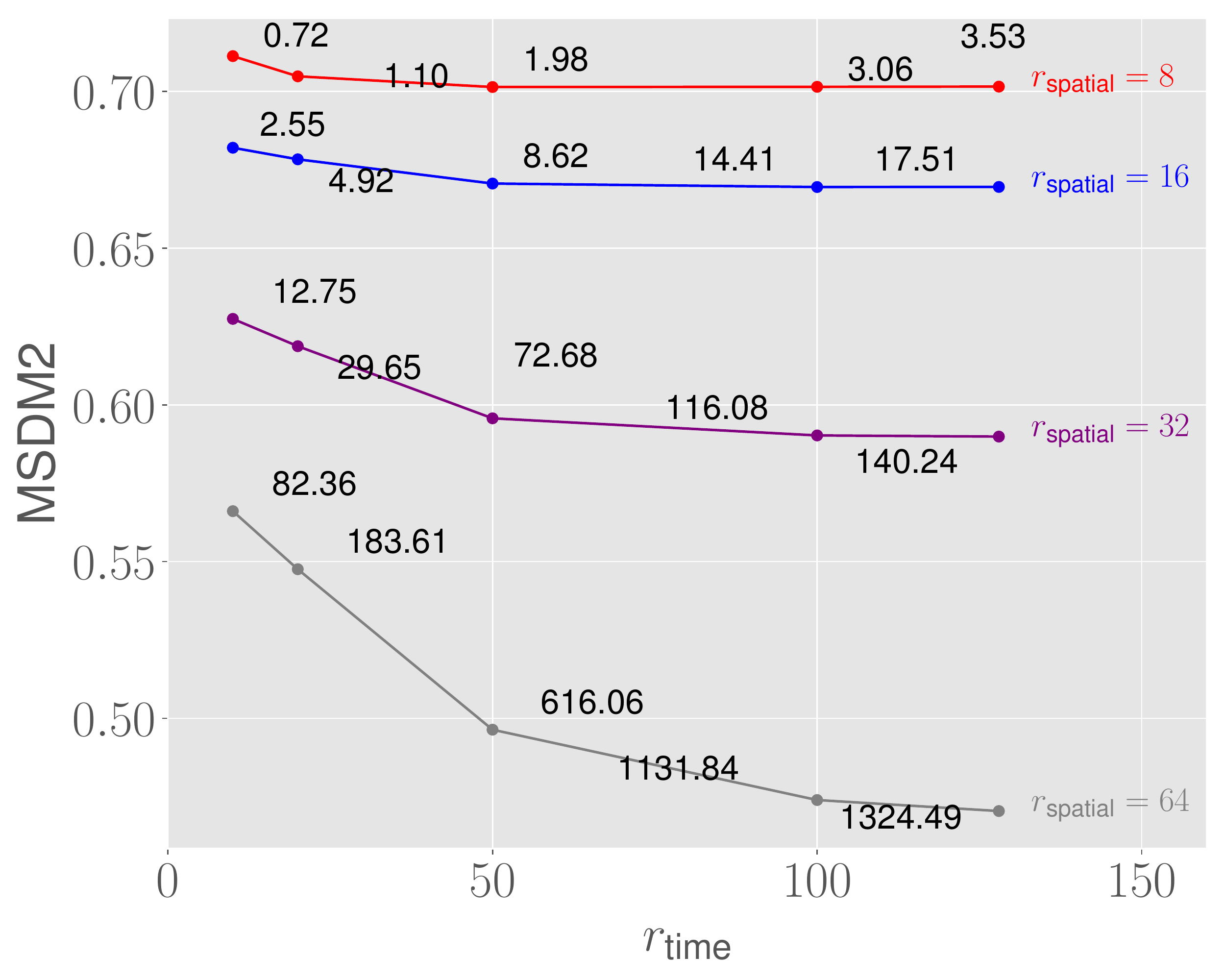}
			\caption{MSDM2}
			\label{fig:metric1}
		\end{subfigure}
		\vspace{0.5em}
		\caption{Ranks vs.\ performance for TT-Tucker compressed \textit{longshort-flying-eagle} scene of resolution $512^3$ with 284 time frames. Metrics are averaged between frames 1, 142, and 284. Each data point is annotated with the corresponding compression ratio scaled by $10^6$.}
		\label{fig:metrics}
	\end{figure}
	
	Note that this procedure is compatible with any format that supports rank truncation, which includes the TT and Tucker formats and mixtures thereof that allow for concatenation in compressed format. They are implemented via concatenation of the cores, increasing the rank of the resultant cores to a sum of the ranks of the operands. Afterward, rank truncation is applied to reduce the rank back to the desired maximal value with an SVD-like algorithm~\cite{oseledets2011tensor}.
	
	\paragraph{QTT scene.}
	The scene stored in QTT format has the shape
	$x_1\times y_1\times z_1 \times t_1 \times x_2\times y_2\times z_2 \times t_2 \times \cdots \times x_k\times y_k\times z_k \times t_k$, where $x_i = y_i = z_i = t_i = 2$ and $\prod_{i=1}^{\ceil{\log_2 I_1}} x_i = W$, $\prod_{i=1}^{\ceil{\log_2 I_2}} y_i = D$, $\prod_{i=1}^{\ceil{\log_2 I_3}} z_i = H$, $\prod_{i=1}^{\ceil{\log_2 I_4}} t_i = T$, with uncompressed scene shape $W\times D \times H \times T$.
	In order to reconstruct the $i$-th frame from a scene compressed in this way, one must first compute the binary representation of $i$ and then decompress the sub-tensor at $[:, :, :, i_{\text{bit}_k}, :, :, :, i_{\text{bit}_{k - 1}}, \cdots ,:, :, :, i_{\text{bit}_{0}}]$.
	
	\paragraph{OQTT scene.}
	The scene stored in OQTT format has the shape
	$x_1 y_1 z_1 \times t_1 \times x_2 y_2 z_2 \times t_2 \times \cdots \times  x_k y_k z_k \times t_k$, where $x_i = y_i = z_i = t_i = 2$ and $\prod_{i=1}^{\ceil{\log_2 I_1}} x_i = W$, $\prod_{i=1}^{\ceil{\log_2 I_2}} y_i = D$, $\prod_{i=1}^{\ceil{\log_2 I_3}} z_i = H$, $\prod_{i=1}^{\ceil{\log_2 I_4}} t_i = T$, with uncompressed scene shape $W\times D \times H \times T$.
	In order to reconstruct the $i$-th frame, one must again first compute the binary representation of $i$, then decompress the sub-tensor at $[:, i_{\text{bit}_k}, :, i_{\text{bit}_{k - 1}}, \cdots ,:, i_{\text{bit}_{0}}]$.
	
	\begin{algorithm}[h!]
		\label{alg:oqtt}
		\begin{algorithmic}
	\STATEx $\sI = I_1 \times I_2 \times I_3$ -- 3D grid 
	\STATEx $\tX \in \R^{\sI}$ -- input 3D volumetric frame
	\STATEx $R$ -- maximal rank
    \STATEx $\displaystyle \tY$ -- Compressed 3D volumetric frame
	\REQUIRE

	\FOR{$\text{d} = 1, 2, 3$}
	    \STATE $\tX \leftarrow \text{pad}(X, 2^{\ceil{\log_2 I_d}})$
	\ENDFOR
	
	\STATE $\tX \leftarrow \text{reshape}(X, 2^{\sum_{d=1}^{3} \ceil{\log_2(I_d)})})$
	\STATE $\tX \leftarrow \text{permute}(X, (x_1, y_1, z_1, \cdots, x_{\ceil{\log_2 I_1}}, y_{\ceil{\log_2 I_2}}, z_{\ceil{\log_2 I_3}}))$,\newline $I_1 = \prod_{i = 1}^{\ceil{\log_2 I_1}} x_i$, $I_2 = \prod_{i = 1}^{\ceil{\log_2 I_2}} y_i$ , $I_3 = \prod_{i = 1}^{\ceil{\log_2 I_3}} z_i$
	\STATE $\tY \leftarrow \text{TT}(\tX, R)$
	
\end{algorithmic}
		\caption{Conversion of a single 3D volumetric frame into OQTT format.}
	\end{algorithm}
	
	\section{Results}
	\label{sec:results}
	\subsection{Data}
	For our experiments, we use selected scenes from the CAPE dataset \cite{CAPE:CVPR:20,pons2017clothcap}.
	The dataset consists of 3D meshes of 15 (10 male, 5 female)
	clothed people in motion. For each scene frame, we compute its TSDF and discretize it at resolution $512^3$, using the PySDF library.
	We use $\tau = 0.05$ in \cref{eq:tsdf} to allow $\approx 10$ voxels with distinct levels near the TSDF zero level set.
	Since all scene frames must share a common coordinates frame, we recover the scene's bounding box and compute the SDF of each frame within that box.
	
	We further demonstrate the performance of \acrotfourdt{} on selected scenes from the Articulated Mesh Animation dataset~\cite{vlasic2008articulated}. That real-world dataset contains 10 mesh sequences depicting 3 different humans performing various actions.
	For each scene frame, we compute its TSDF and discretize it at resolution $512^3$, using the PySDF library.
	We use $\tau = 0.05$ in \cref{eq:tsdf} to allow $\approx 10$ voxels with distinct levels near the TSDF zero level set.
	
	\subsection{Influence of Tensor Ranks}
	We provide several error metrics computed for the TT-Tucker format, averaged across the first, middle, and last frames of the \textit{longshort-flying-eagle} scene, for different spatial and temporal ranks in \cref{fig:metrics}.
	See \cref{subsec:metrics} for the definition of the error metrics.
	Note that the time dimension is far from full-rank in the TT-Tucker format since the error metrics saturate well below the total sequence length of 284 frames.
	
	The best compression/performance was obtained for the OQTT base format.
	Selected visualizations are provided in \cref{sec:visualisations}: \cref{fig:tt_metrics} shows error metrics for the TT format, as a function of the rank. Qualitative results are shown in \cref{fig:compression_results} for the TT-Tucker format, in \cref{fig:tt_compression_results} for the TT format, and in \cref{fig:compression_qtt} for QTT. For visual results of OQTT please refer to \cref{fig:compression_oqtt}.
	In the cases of TT and TT-Tucker, severe rank truncation smoothes the reconstructed surface and erodes smaller features like fingers or facial details. In contrast, QTT preserves more details at the cost of discontinuities due to the separation and independent compression of octet sub-dimensions.
	OQTT reduces these discontinuity artifacts by encoding the octets as a single dimension without compressing the ranks between sub-dimensions of a single octet. See \cref{fig:compression_oqtt} for qualitative results for OQTT.
	We also present quantitative OQTT compression results in \cref{tab:metrics}, at a rank that offers a good compromise between visual quality and compression rate.
	
	\begin{table*}[h!]
		\centering
		\small
		\setlength\tabcolsep{2.pt}
		\begin{tabular}{ ccccc }
	\toprule
	&
	\begin{minipage}{0.18\linewidth}
		\centering
		Crane\\
		\fixedvspace{0.4\baselineskip}
		\resizebox{\linewidth}{!}{
			\includegraphics[trim=170 200 200 170, clip]{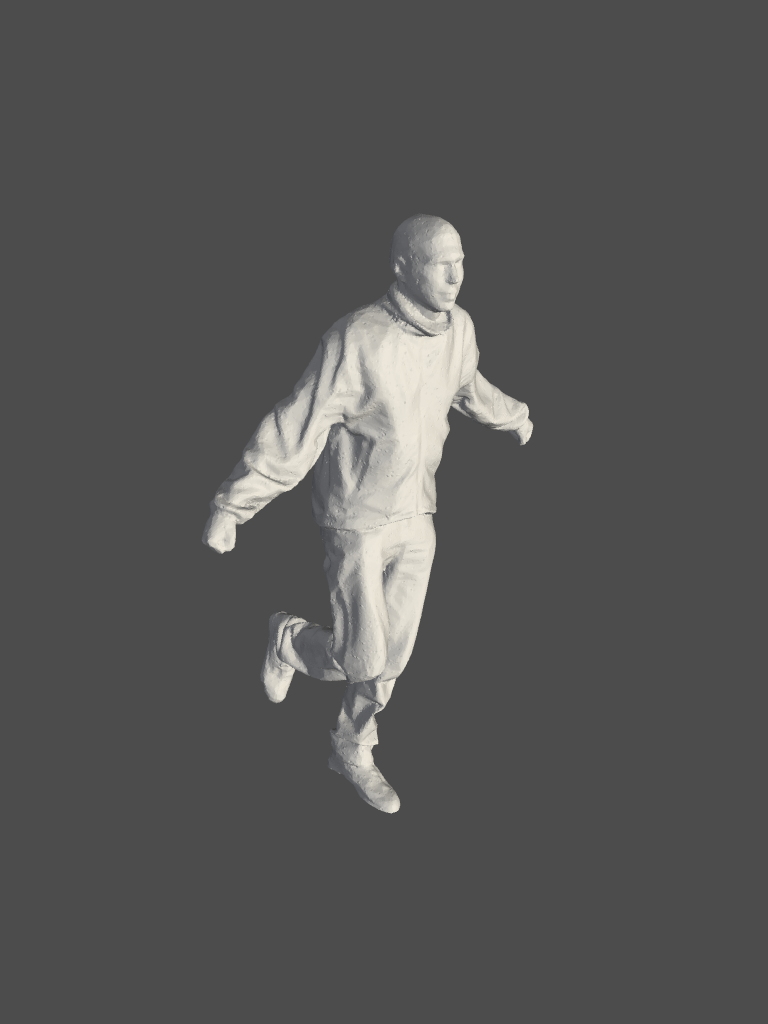}
		}
	\end{minipage} &
	\begin{minipage}{0.18\linewidth}
		\centering
		Swing\\
		\fixedvspace{0.4\baselineskip}
		\resizebox{\linewidth}{!}{
			\includegraphics[trim=170 200 200 170, clip]{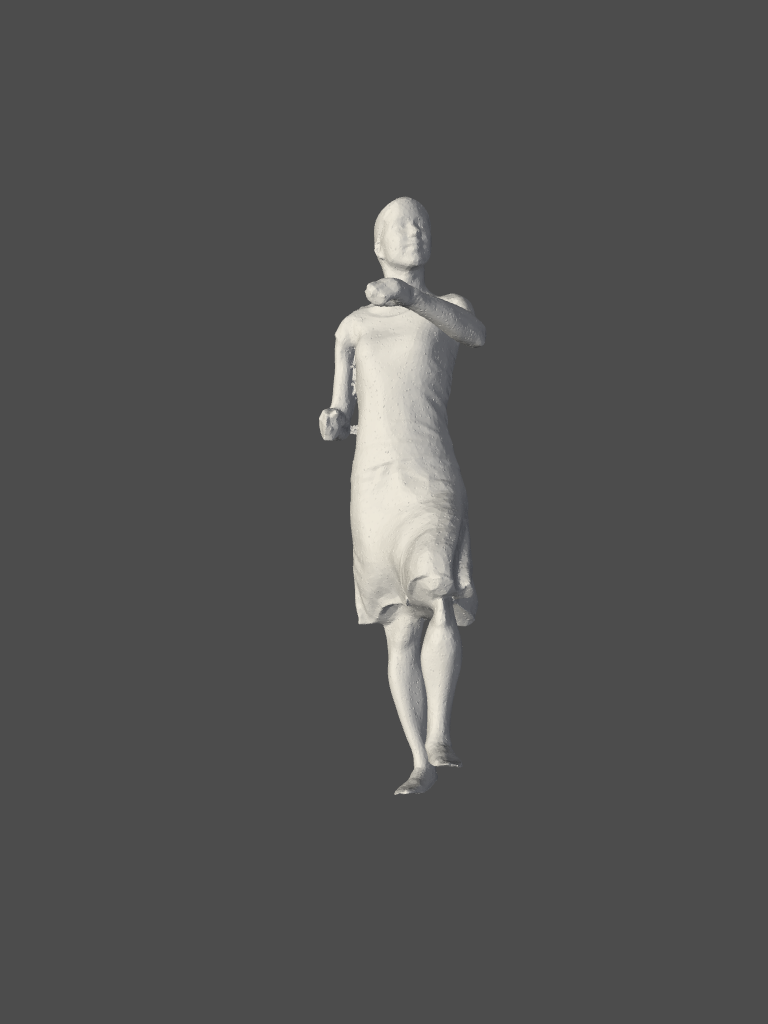}
		}
	\end{minipage} &
	\begin{minipage}{0.18\linewidth}
		\centering
		Handstand\\
		\fixedvspace{0.4\baselineskip}
		\resizebox{\linewidth}{!}{
			\includegraphics[trim=170 200 200 170, clip]{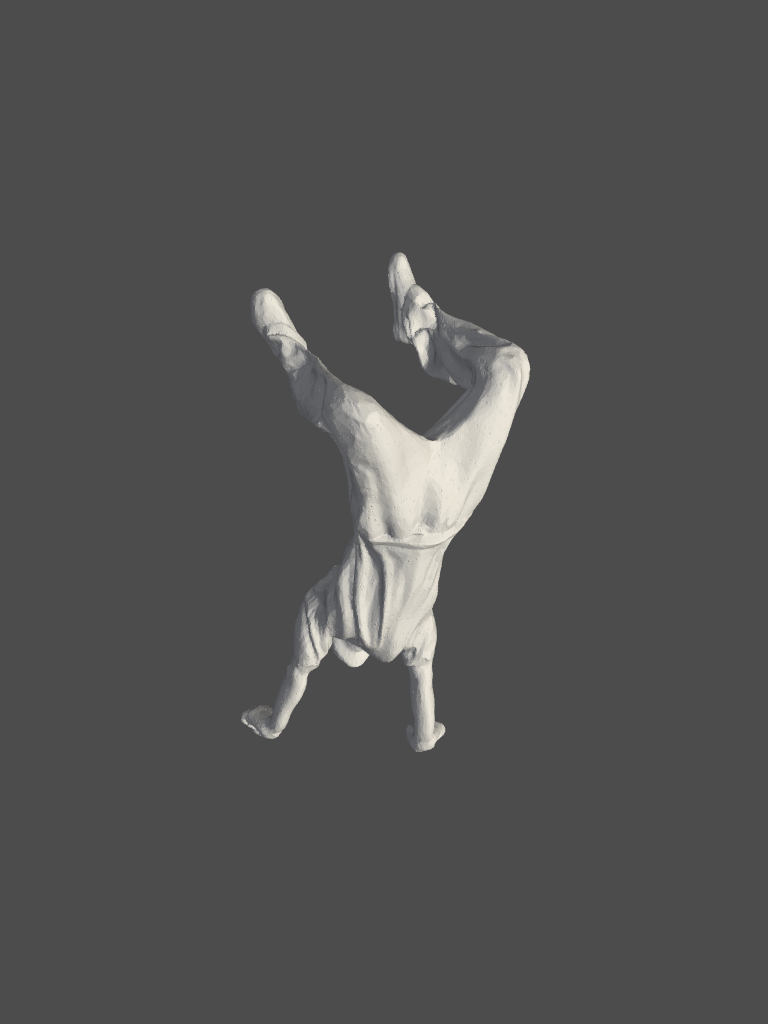}
		}
	\end{minipage} &
	\begin{minipage}{0.18\linewidth}
		\centering
		Samba\\
		\fixedvspace{0.4\baselineskip}
		\resizebox{\linewidth}{!}{
			\includegraphics[trim=170 200 200 170, clip]{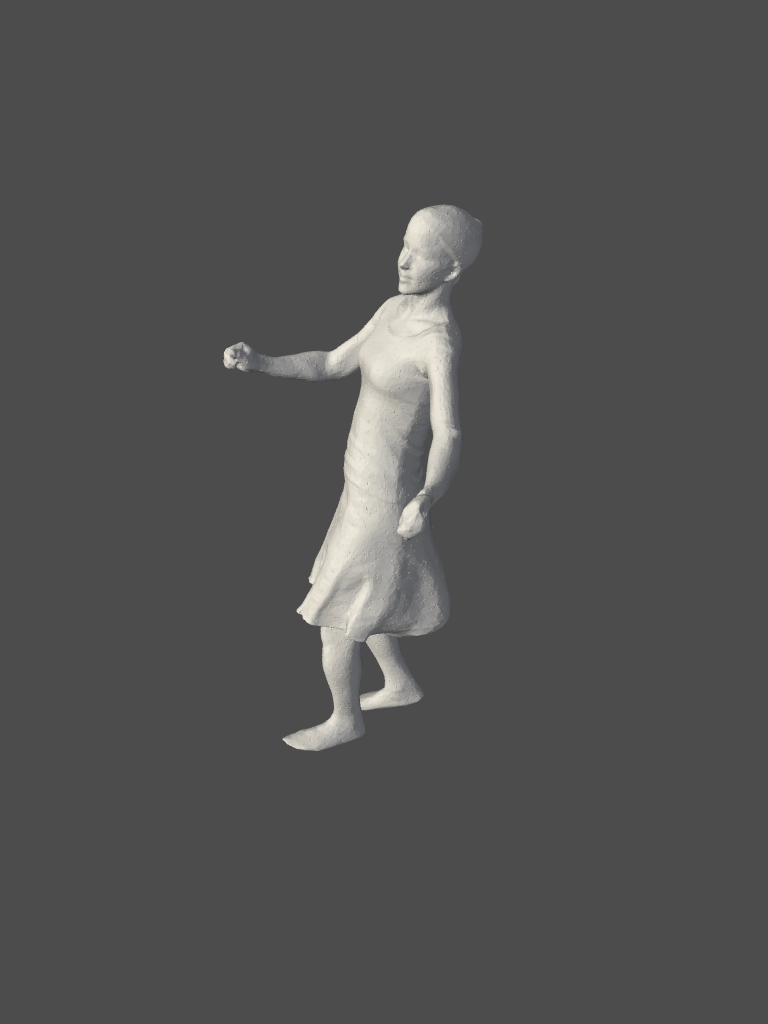}
		}
	\end{minipage} \\
	\\
	Resolution & $512^3\times174$ & $512^3\times174$ & $512^3\times174$ & $512^3\times149$\\
	\\
	Metric \\
	\midrule
	\rowcolor{gray!12}
	L2 $\downarrow$                 & 2.67                 & 2.07                 & 2.45                 & 1.71 \\
	Chamfer distance   $\downarrow$ & $5e^{-5}$            & $4e^{-5}$            & $5e^{-5}$            & $4e^{-5}$ \\
	\rowcolor{gray!12}
	Hausdorff distance $\downarrow$ & 0.190                 & 0.015                & 0.012                & 0.013 \\
	MSDM2 $\downarrow$              & 0.36                 & 0.38                 & 0.35                 & 0.36 \\
	\rowcolor{gray!12}
	IoU $\uparrow$                  & 0.41                 & 0.40                  & 0.64                 & 0.54 \\
	Compression                     & 1:954                & 1:1356               & 1:938                & 1:935 \\
	\rowcolor{gray!12}
	Size                            & \SI{549}{\mega\byte} & \SI{386}{\mega\byte} & \SI{558}{\mega\byte} & \SI{560}{\mega\byte} \\
	\bottomrule
\end{tabular}
		\vspace{1em}
		\caption{Quantitative performance of OQTT for selected scenes from the Articulated Mesh Animation dataset~\cite{vlasic2008articulated}. The metrics are averaged across the first, middle, and last frames. For all scenes, $r_{\max} = 4000$, chosen to provide a good trade-off between visual quality and compression rate.}
		\label{tab:metrics}
	\end{table*}
	
	\section{Failure Cases and Limitations}
	\label{sec:limitations}
	We did not observe significant failures while working with the CAPE and AMA datasets. However, how \acrotfourdt{} fares for 3D data that are not structurally sparse (i.e., not mostly free space) remains to be tested. Unfortunately, there are no such scenes in the two datasets.
	
	We do note that TT decomposition is not invariant against the rotation of the input. Hence, one could attack the scheme by rotating the scene w.r.t.\ the grid axes in a way that maximizes the reconstruction error. A straightforward measure to mitigate the influence of the grid orientation is to pre-rotate the 3D scene to its principal axes before grid sampling and OQTT compression.
	
	\section{Conclusions}
	\label{sec:conclusions}
	We have presented \acrotfourdt{}, a scalable and interpretable compression pipeline for 3D time sequence data based on tensor decomposition.
	Our scheme improves over the related TT-TSDF method in two ways:
	(i) We are able to process temporally varying data. We can do so even though the TSDF field takes hundreds of GBs, and our method works fully in-memory;
	(ii) We tested various tensorization schemes, such as a TT-Tucker hybrid, the QTT format, and the new OQTT variant, and found OQTT to outperform previous decompositions.
	OQTT is tailored specifically to 3D volumetric scene representations, and our experiments quantitatively and qualitatively support its special reshaping of the 3D grid.
	
	\clearpage
	
	\bibliographystyle{apalike}
	\bibliography{bibliography}
	
	\appendix
	\clearpage
	\section{Appendix}
\label{sec:appendix}

\subsection{Metrics}
\label{subsec:metrics}

Throughout this work, we convert temporal 3D data from mesh format to tensor format for compression using sampled truncated SDF and from tensor format back to temporal 3D data in the form of meshes for quantitative tests using the marching cubes algorithm.

We selected a tensor-based l2 reconstruction metric for the comparison in tensorial form.
We use the metrics below to evaluate the quality of the obtained mesh.

\subsubsection{Intersection over Union (IoU)}
IoU is a commonly used metric for comparing 3D shapes~\cite{rezatofighi2019generalized}.
For two given 3D volumes $\tA$ and $\tB$:
\begin{equation}
	\text{IoU}(\tA, \tB) = \frac{\|\tA \cap \tB\|}{\|\tA \cup  \tB\|}.
\end{equation}
3D mesh is converted to an occupancy grid, and the operation is easily performed with implementation from \cite{KaolinLibrary}.

\subsubsection{Hausdorff distance}
One-sided Hausdorff distance is computed as:
\begin{equation}
	d_\text{h}(\tA, \tB) = \max \{\min\{ ||a - b||: a \in \tA\} : b \in \tB\}.
\end{equation}
Note, that $d_\text{h}$ is not symmetric.
The symmetric version is defined as :
\begin{equation}
	d_\text{H}(\tA, \tB) = \max \{d_\text{h}(\tA, \tB), d_\text{h}(\tB, \tA)\}.
\end{equation}
We use the implementation from \cite{libigl}.

\subsubsection{Chamfer distance}
Symmetric Chamfer distance is defined as:
\begin{equation}
	d_\text{CD}(\tA, \tB) = \sum_{a \in \tA} \min_{b \in \tB} ||x - y||^2 + \sum_{b \in \tB} \min_{a \in \tA} ||x - y||^2.
\end{equation}
Following \cite{boyko2020tt}, we sample 30'000 points from each mesh and compute sampled symmetric Chamfer distance. Finally, we use the implementation from \cite{KaolinLibrary}.

\subsubsection{Mesh Structural Distortion Measure (MSDM2)}
Introduced in \cite{lavoue2011multiscale} MSDM2 metric is used to estimate the correlation with human visual perception.
The metric assumes one mesh to be original and the second distorted.
See \cref{alg:msdm2} for an algorithmic view of MSDM2 computation.
\begin{algorithm}[h!]
	\caption{MSDM2 high-level computation scheme}
	\label{alg:msdm2}
	\begin{algorithmic}
    \STATEx $M_o$ -- original mesh, $M_d$ -- distorted mesh, $\{v_i \in M\}$ -- set of mesh vertices
	\REQUIRE
    \FOR{$v_i \in M_o$}
    \STATE Find $v_j = \min_{v_k \in M_d} \text{dist}(v_i, v_k)$
    \ENDFOR
    \FOR{$v_i \in M_o$}
    \STATE calculate curvature of $v_i$
    \ENDFOR
    \FOR{$v_i \in M_d$}
    \STATE calculate curvature of $v_i$
    \ENDFOR
    \STATE interpolate curvature per face for $M_o$ and $M_d$
    \FOR{$v_i \in M_d$}
    \STATE calculate MSDM2 for $v_i$ based on curvature features and nearest neighbours $v_j$ from $M_o$ 
    \ENDFOR
    \STATE Integrate global asymmetric MSDM2
\end{algorithmic}

\end{algorithm}
Refer to \cite{lavoue2006perceptually,lavoue2011multiscale} for the more details on metric formulation.
We use original implementation from \cite{lavoue2011multiscale} and self written pybind11 \cite{pybind11} interface to Python.

\clearpage
\subsection{Selected Visualizations}
\label{sec:visualisations}

\begin{figure}[h!]
	\centering
	\begin{subfigure}[b]{0.49\textwidth}
		\includegraphics[width=\linewidth]{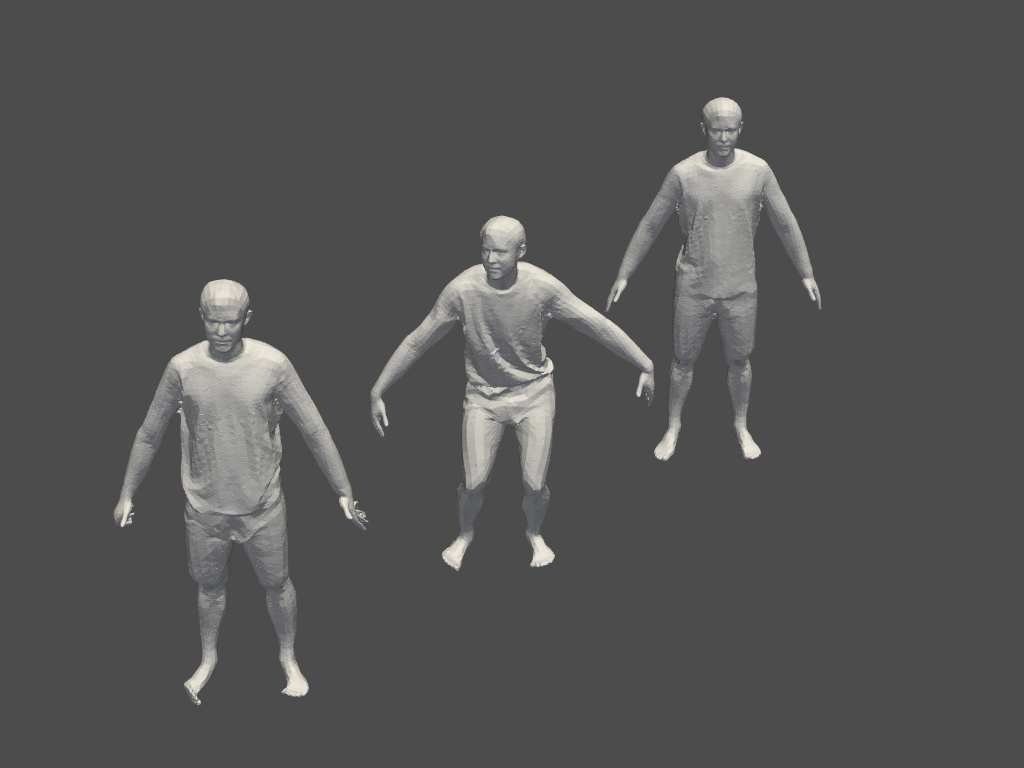}
		\caption{Original, \SI{284}{\giga\byte}}
		\label{fig:original}
	\end{subfigure}
	\hfill
	\begin{subfigure}[b]{0.49\textwidth}
		\includegraphics[width=\linewidth]{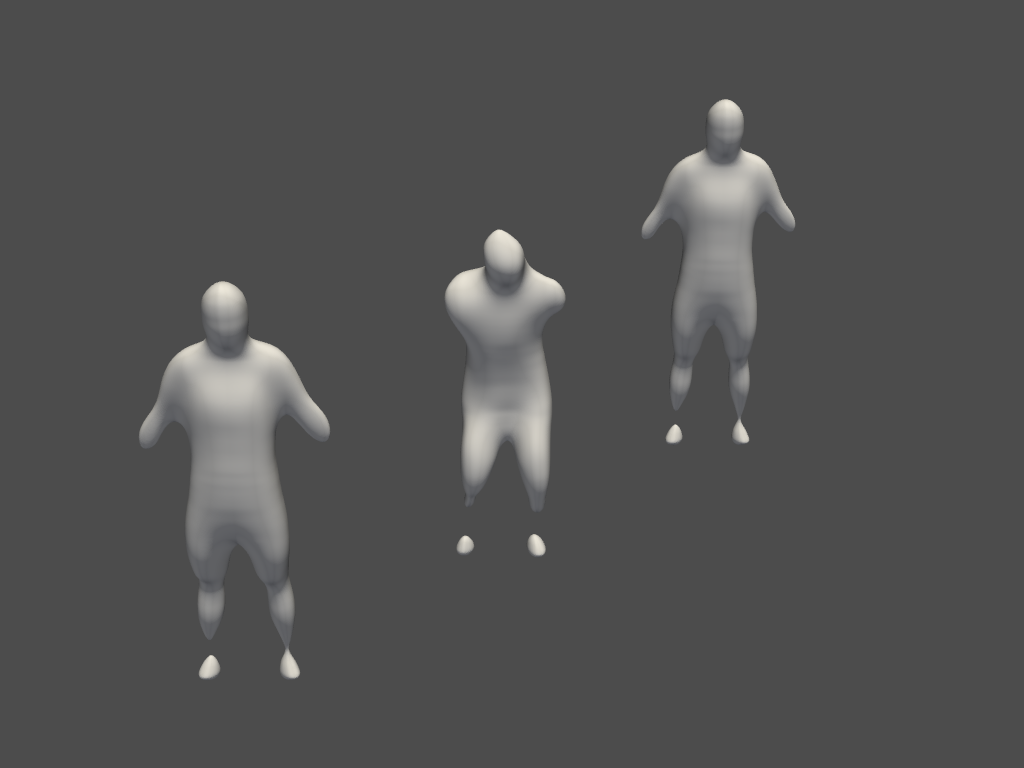}
		\caption{Compression 1:769230, \SI{0.2}{\mega\byte}}
		\label{fig:compressed1}
	\end{subfigure}
	\\
	\begin{subfigure}[b]{0.49\textwidth}
		\includegraphics[width=\linewidth]{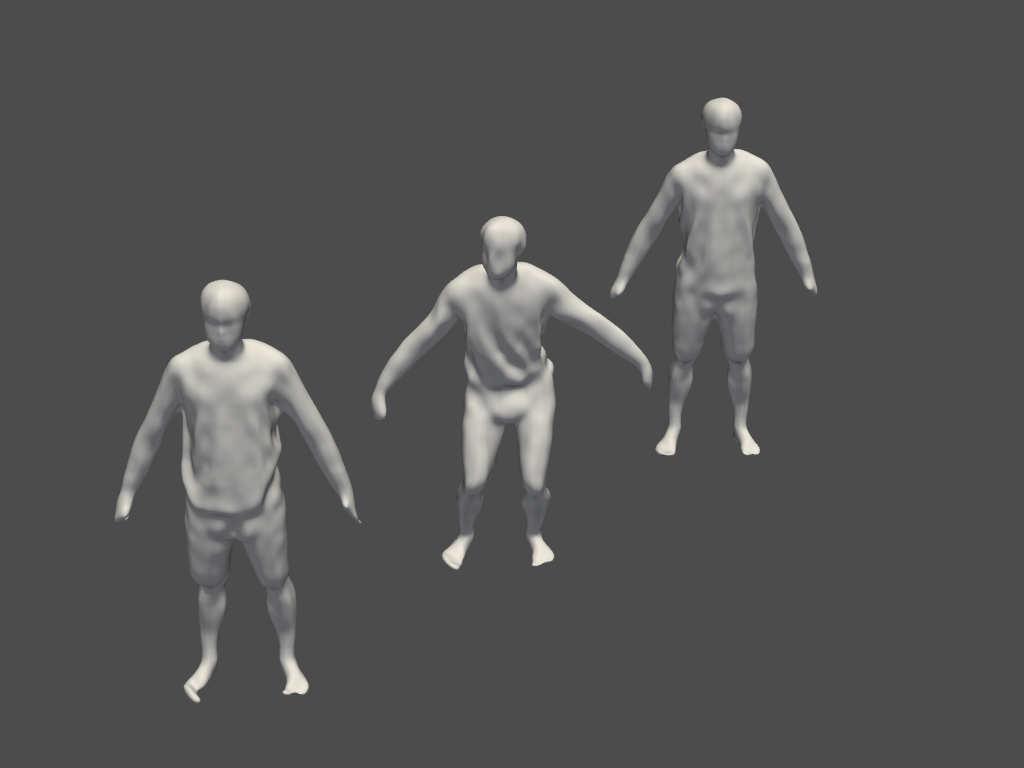}
		\caption{Compression 1:1333, \SI{33}{\mega\byte}}
		\label{fig:compressed2}
	\end{subfigure}
	\hfill
	\begin{subfigure}[b]{0.49
			\textwidth}
		\includegraphics[width=\linewidth]{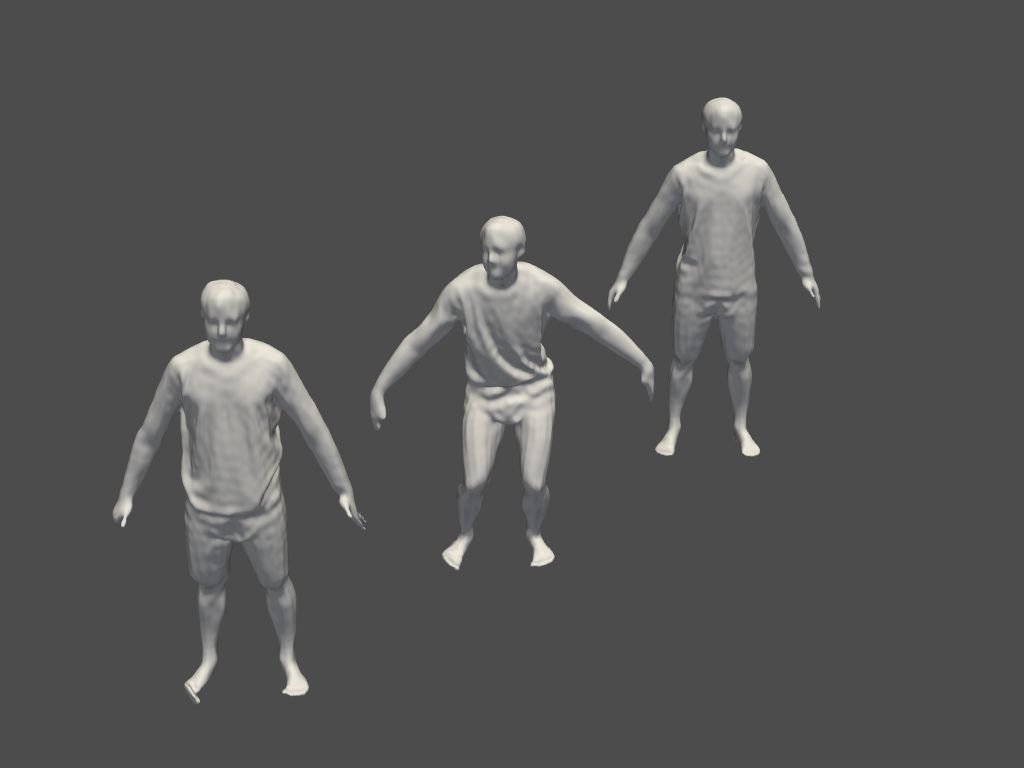}
		\caption{Compression 1:116, \SI{385}{\mega\byte}}
		\label{fig:compressed3}
	\end{subfigure}
	\caption{\acrotfourdt{} with different compression levels in TT-Tucker format for a \textit{longshort-flying-eagle} scene of resolution $512^3$ with 284 time frames. Only frames 1, 142, and 284 are depicted.}
	\label{fig:compression_results}
\end{figure}

\begin{figure}[h!]
	\centering
	\begin{subfigure}[b]{0.49\textwidth}
		\includegraphics[width=\linewidth]{original}
		\caption{Original, \SI{284}{\giga\byte}}
		\label{fig:tt_original}
	\end{subfigure}
	\hfill
	\begin{subfigure}[b]{0.49\textwidth}
		\includegraphics[width=\linewidth]{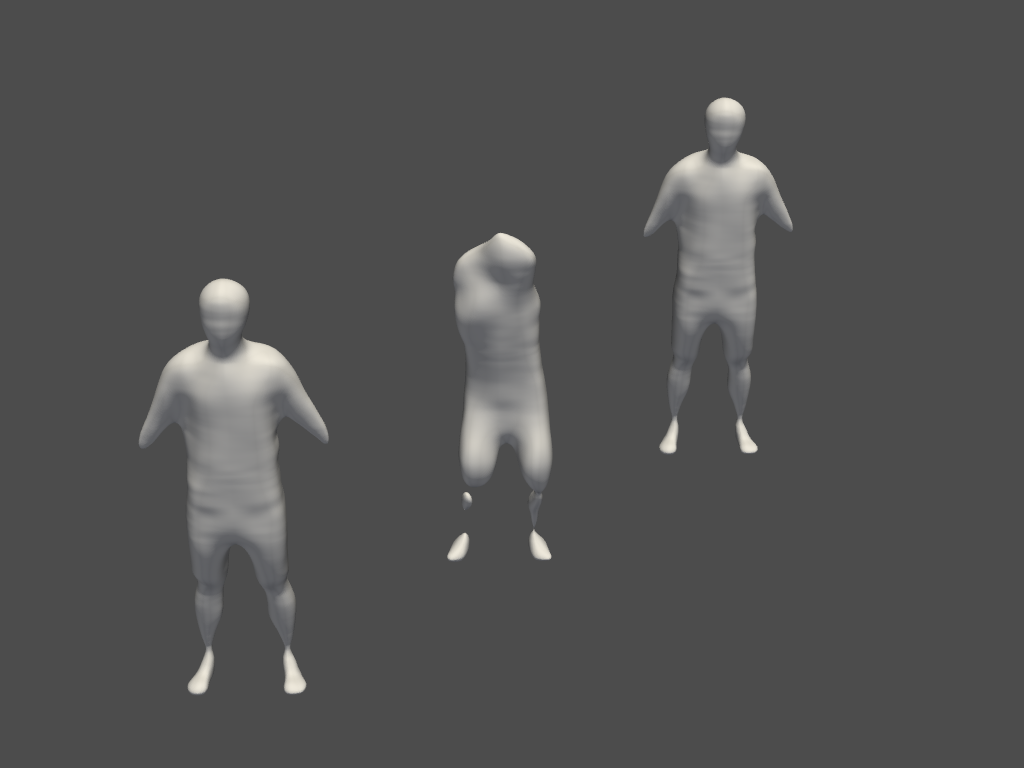}
		\caption{compression 1:476190, \SI{0.61}{\mega\byte}}
		\label{fig:compressed_tt1}
	\end{subfigure}
	\\
	\begin{subfigure}[b]{0.49\textwidth}
		\includegraphics[width=\linewidth]{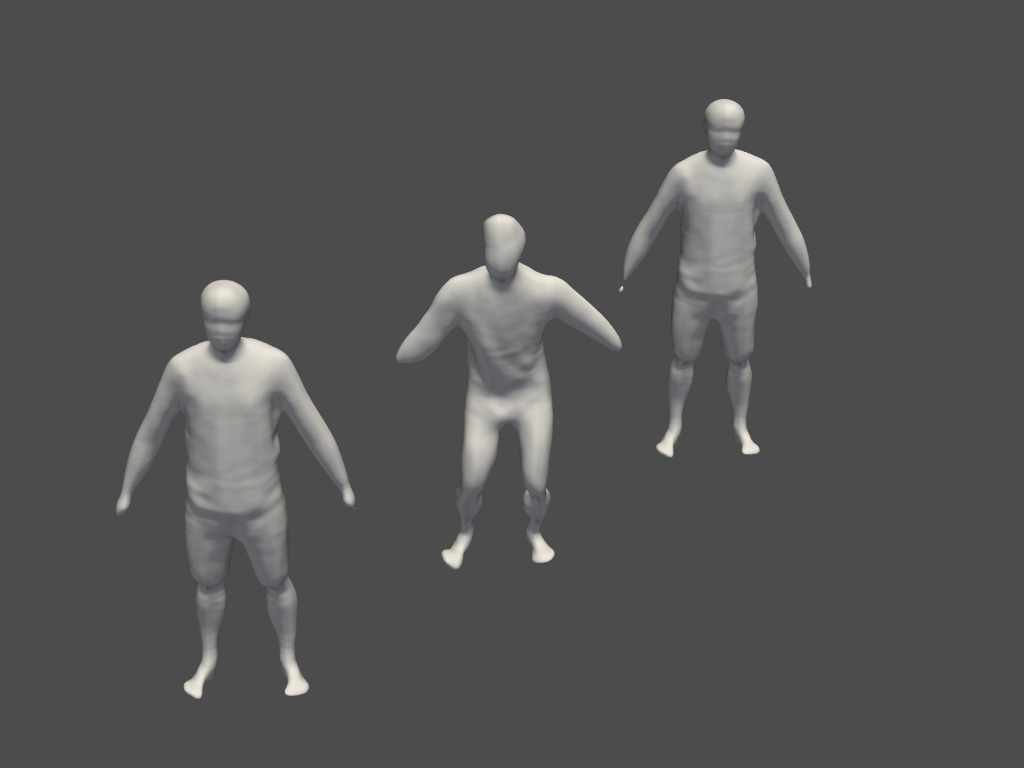}
		\caption{compression 1:6250, \SI{16}{\mega\byte}}
		\label{fig:compressed_tt2}
	\end{subfigure}
	\hfill
	\begin{subfigure}[b]{0.49
			\textwidth}
		\includegraphics[width=\linewidth]{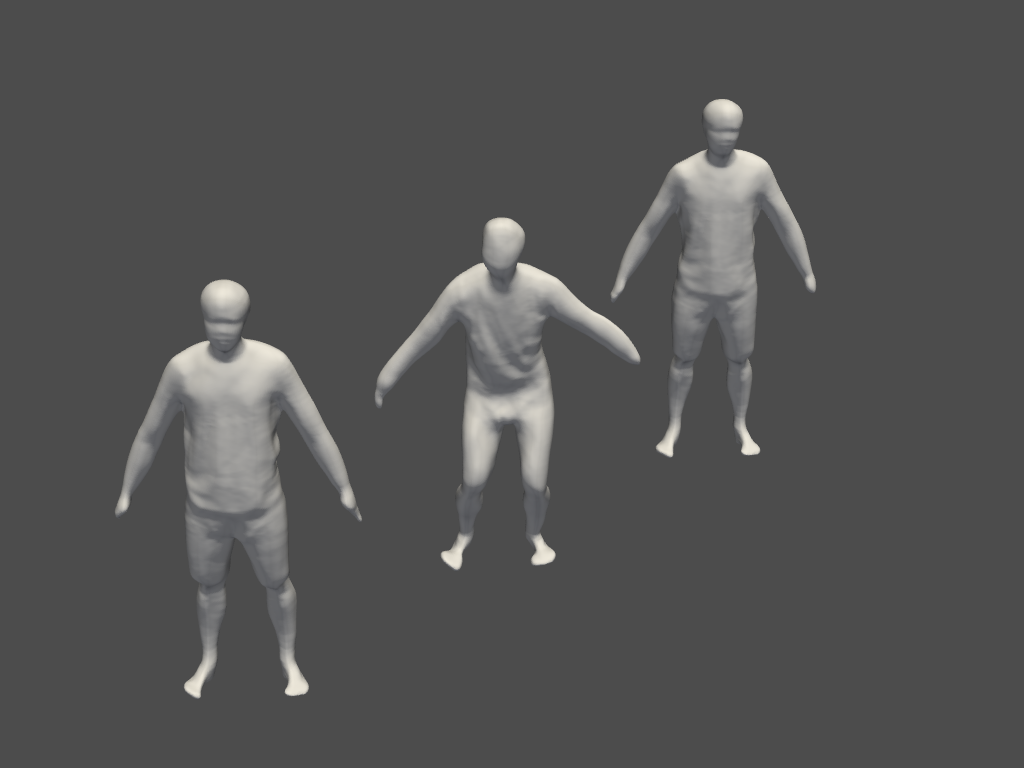}
		\caption{compression 1:2000, \SI{48.5}{\mega\byte}}
		\label{fig:compressed_tt3}
	\end{subfigure}
	\caption{\acrotfourdt{} with different compression levels in TT format for a \textit{longshort-flying-eagle} scene of resolution $512^3$ with 284 time frames. Only frames 1, 142, and 284 are depicted. The compression ratio is different from the actual memory consumption due to the padding of the time dimension to 512.}
	\label{fig:tt_compression_results}
\end{figure}

\begin{figure}[h!]
	\centering
	\begin{subfigure}[b]{0.49\textwidth}
		\includegraphics[width=\linewidth]{original}
		\caption{Original, \SI{284}{\giga\byte}}
		\label{fig:original_qtt}
	\end{subfigure}
	\hfill
	\begin{subfigure}[b]{0.49\textwidth}
		\includegraphics[width=\linewidth]{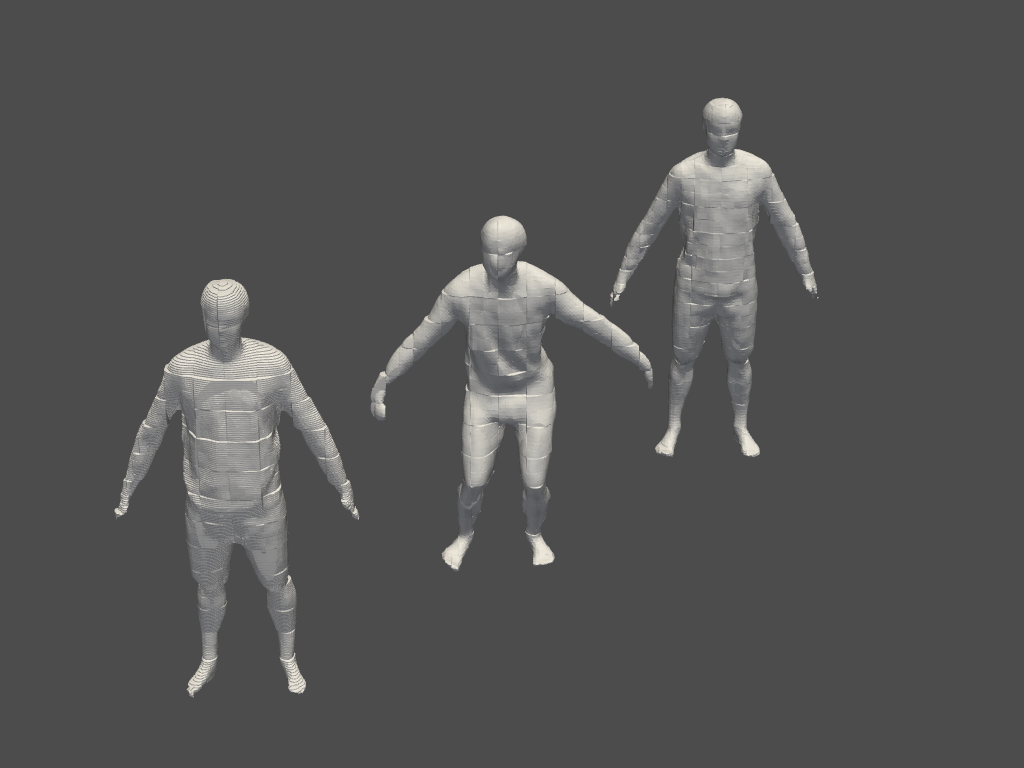}
		\caption{compression 1:6666, \SI{0.08}{\mega\byte}}
		\label{fig:compressed_qtt1}
	\end{subfigure}
	\\
	\begin{subfigure}[b]{0.49\textwidth}
		\includegraphics[width=\linewidth]{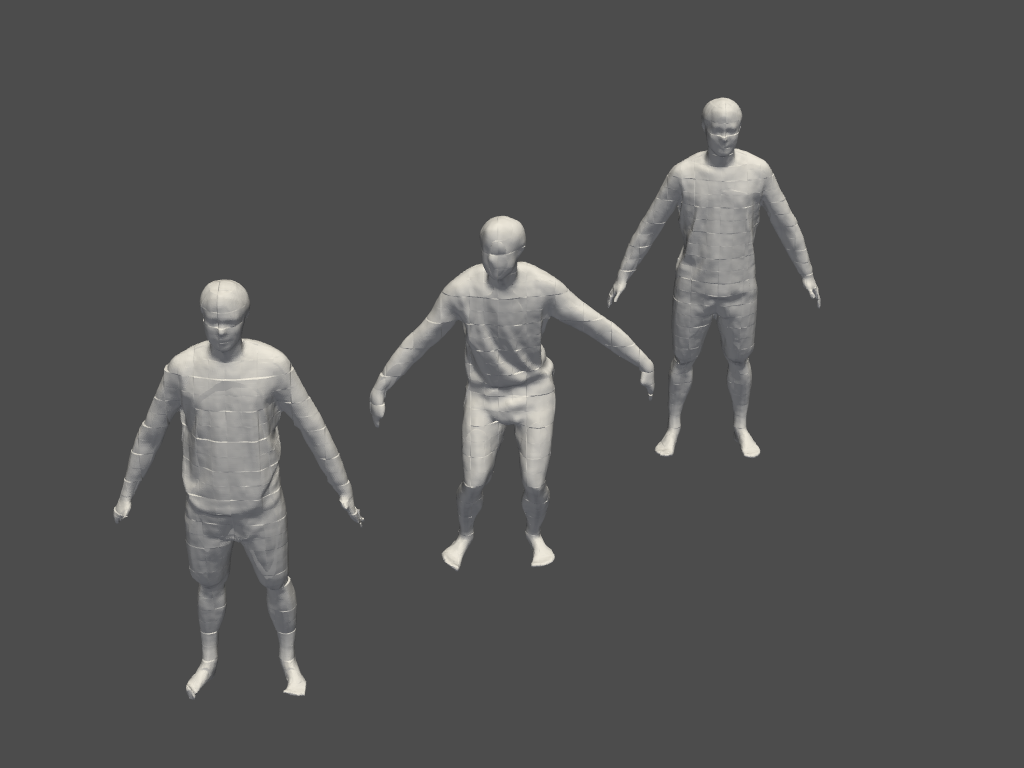}
		\caption{compression 1:1587, \SI{332}{\mega\byte}}
		\label{fig:compressed_qtt2}
	\end{subfigure}
	\hfill
	\begin{subfigure}[b]{0.49
			\textwidth}
		\includegraphics[width=\linewidth]{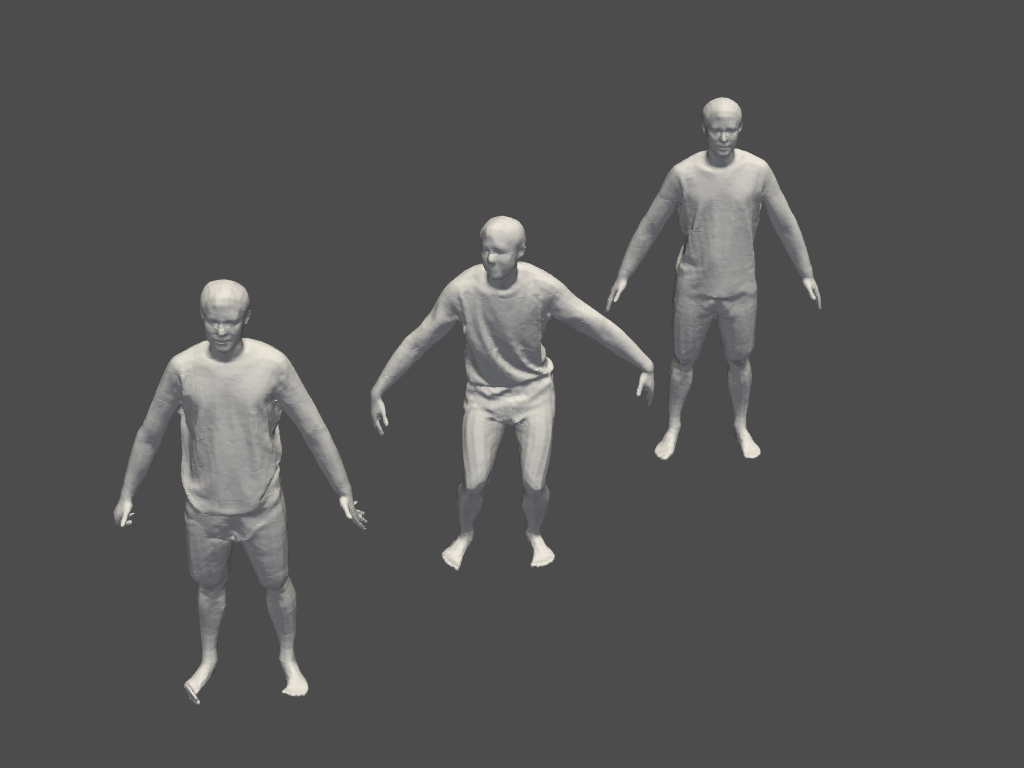}
		\caption{compression 1:244, \SI{1.4}{\giga\byte}}
		\label{fig:compressed_qtt3}
	\end{subfigure}
	\caption{Different levels of compression in QTT format for a \textit{longshort-flying-eagle} scene of resolution $512^3$ with 284 time frames. Only frames 1, 142, and 284 are depicted. The compression ratio is different from the actual memory consumption due to the padding of the time dimension to 512.}
	\label{fig:compression_qtt}
\end{figure}

\begin{figure}[h!]
	\centering
	\begin{subfigure}[b]{0.49\textwidth}
		\includegraphics[width=\linewidth]{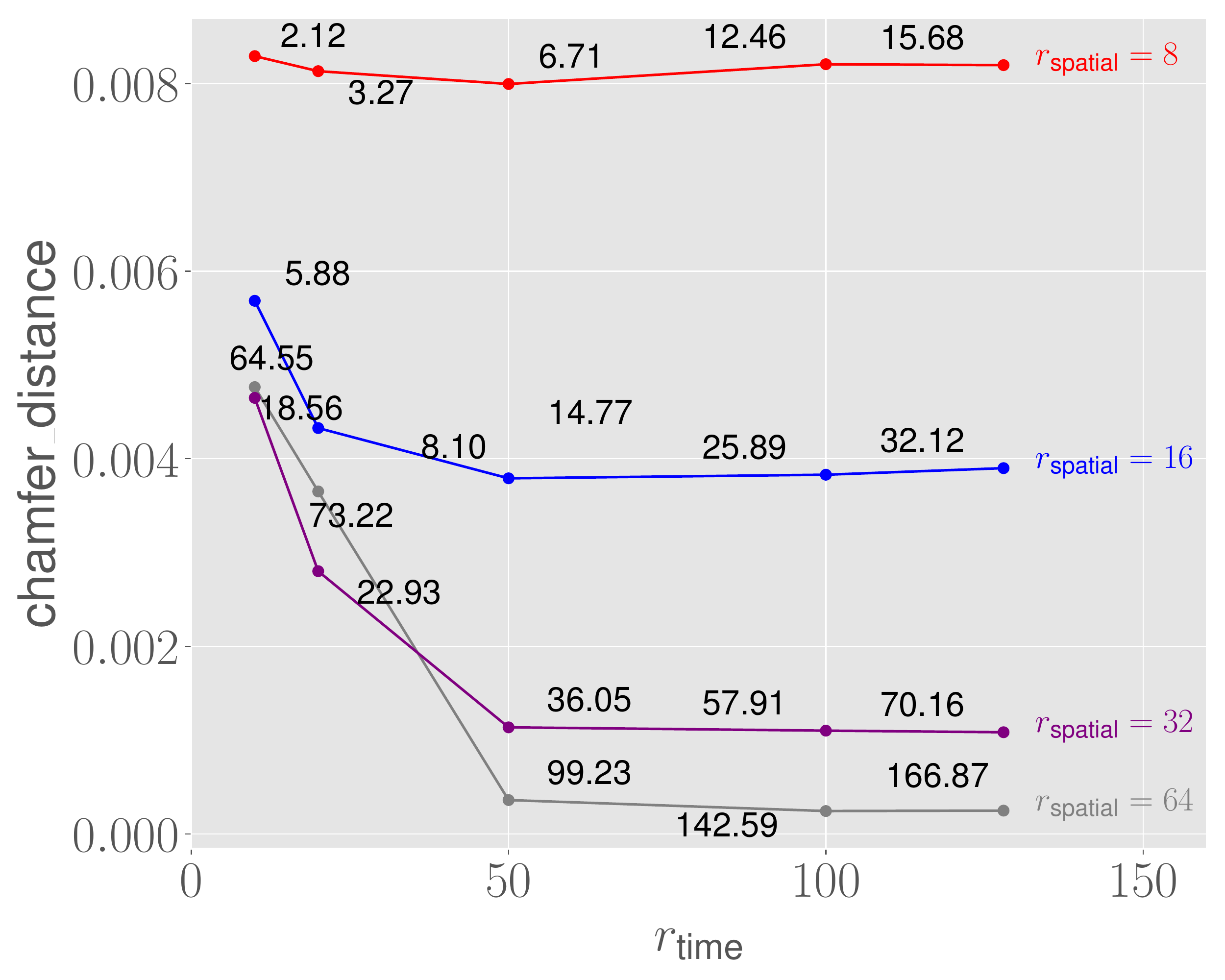}
		\caption{Chamfer distance}
		\label{fig:tt_metric4}
	\end{subfigure}
	\begin{subfigure}[b]{0.49\textwidth}
		\includegraphics[width=\linewidth]{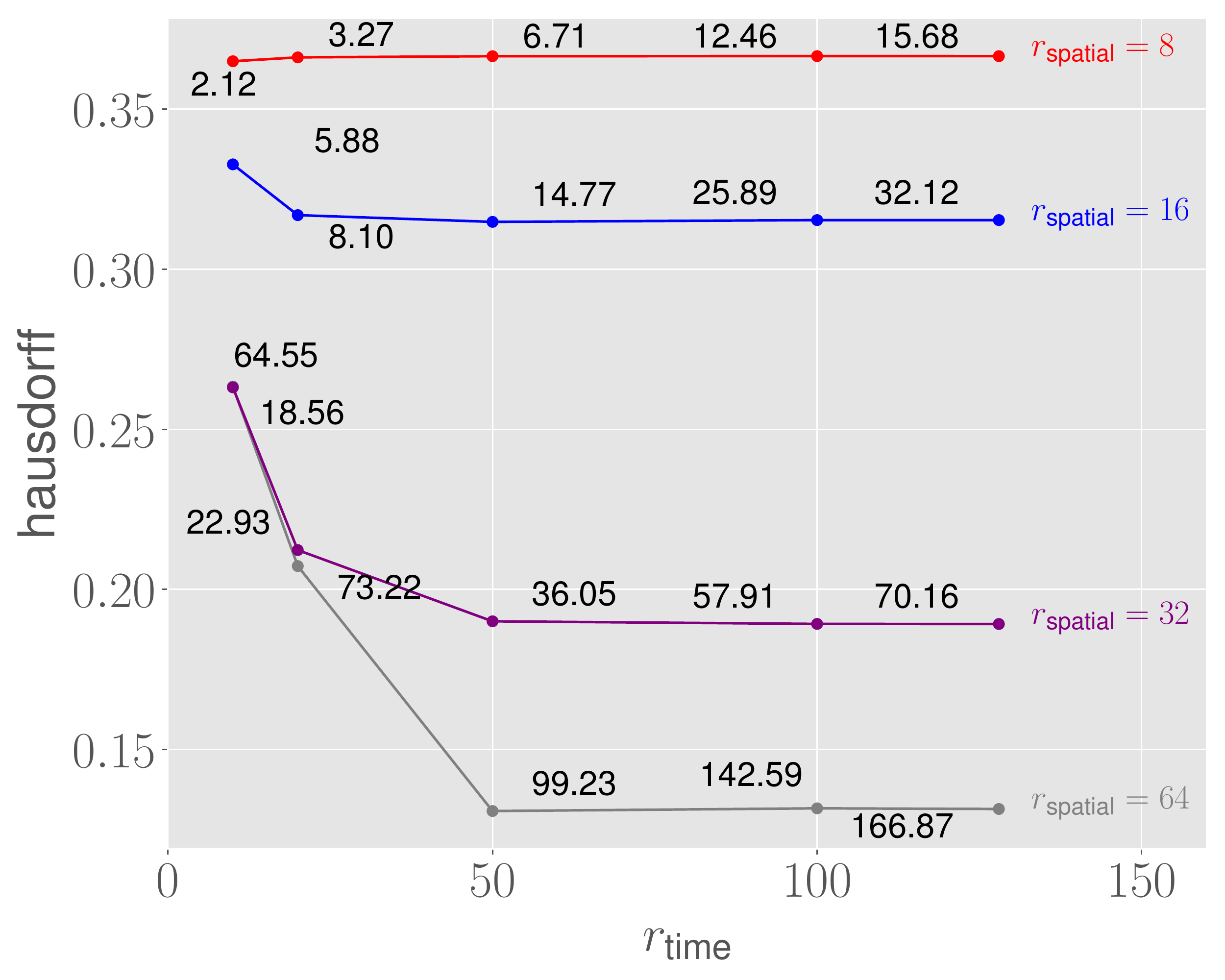}
		\caption{Hausdorff distance}
		\label{fig:tt_metric3}
	\end{subfigure}
	\begin{subfigure}[b]{0.49\textwidth}
		\includegraphics[width=\linewidth]{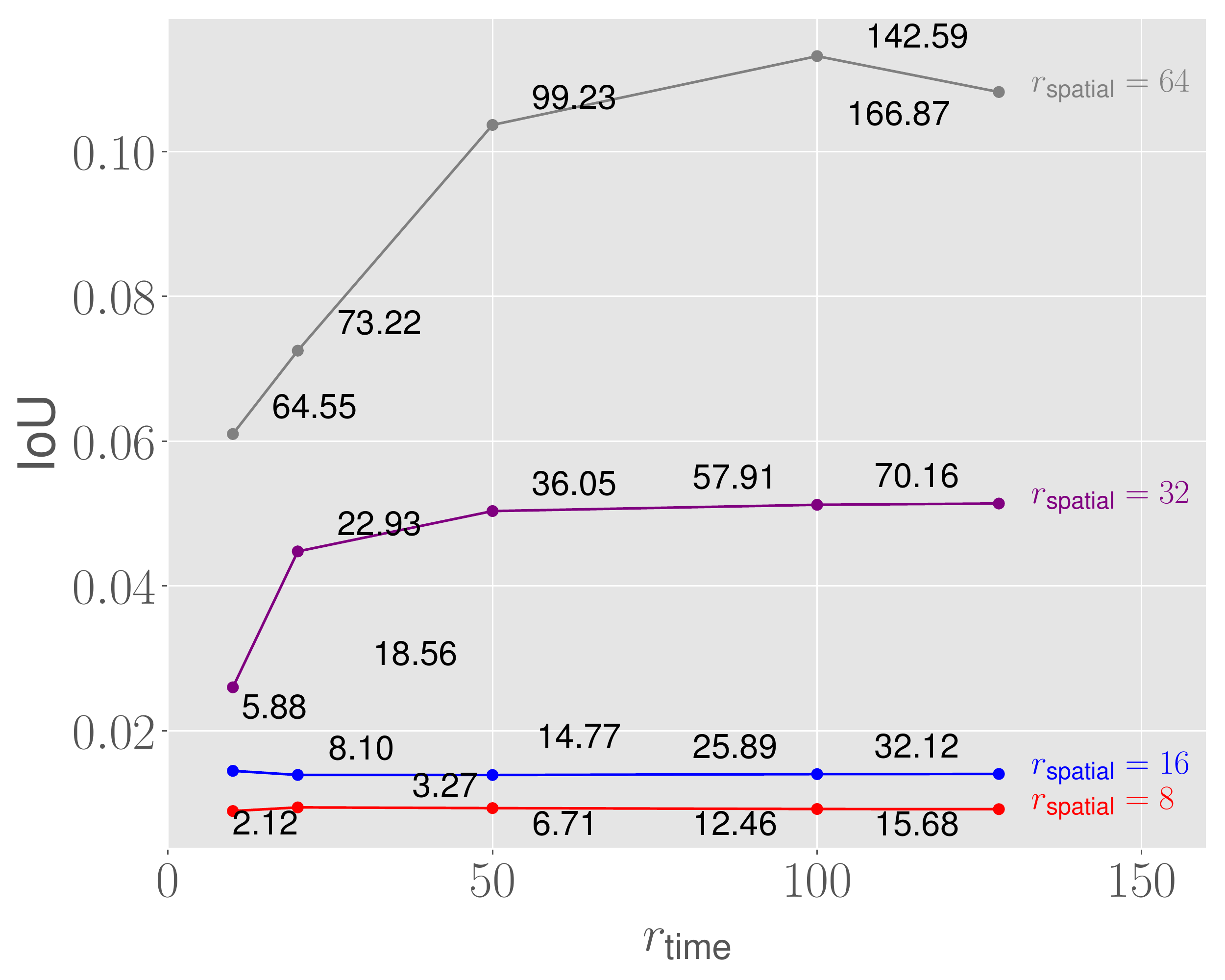}
		\caption{IoU}
		\label{fig:tt_metric2}
	\end{subfigure}
	\begin{subfigure}[b]{0.49\textwidth}
		\includegraphics[width=\linewidth]{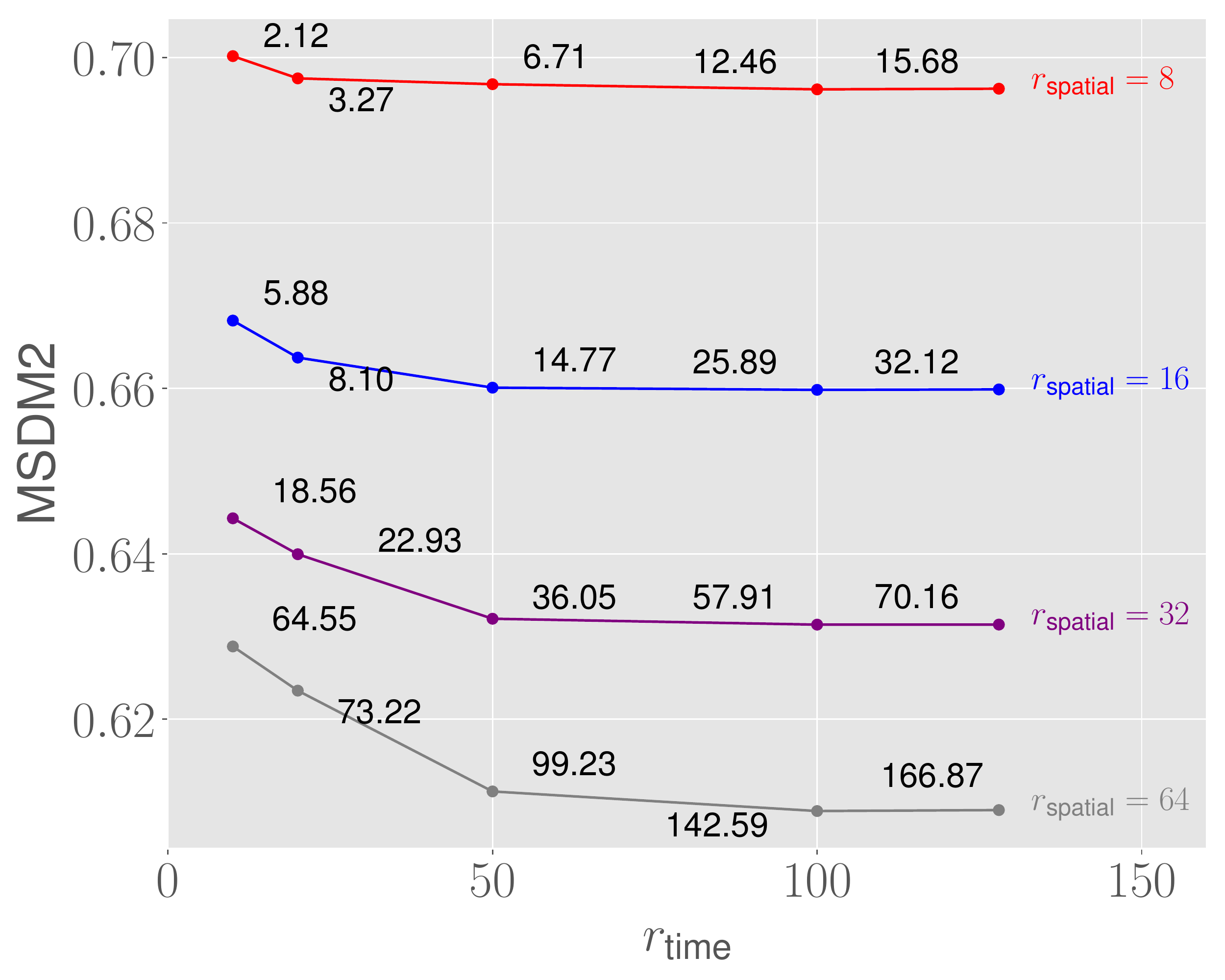}
		\caption{MSDM2}
		\label{fig:tt_metric1}
	\end{subfigure}
	\caption{Ranks vs. performance for TT compressed \textit{longshort-flying-eagle} scene of resolution $512^3$ with 284 time frames. Metrics are averaged between frames 1, 142, and 284. Each data point is annotated with the corresponding compression ratio scaled with 1e6.}
	\label{fig:tt_metrics}
\end{figure}

\begin{figure}[h!]
	\centering
	\begin{subfigure}[b]{0.49\textwidth}
		\includegraphics[width=\linewidth]{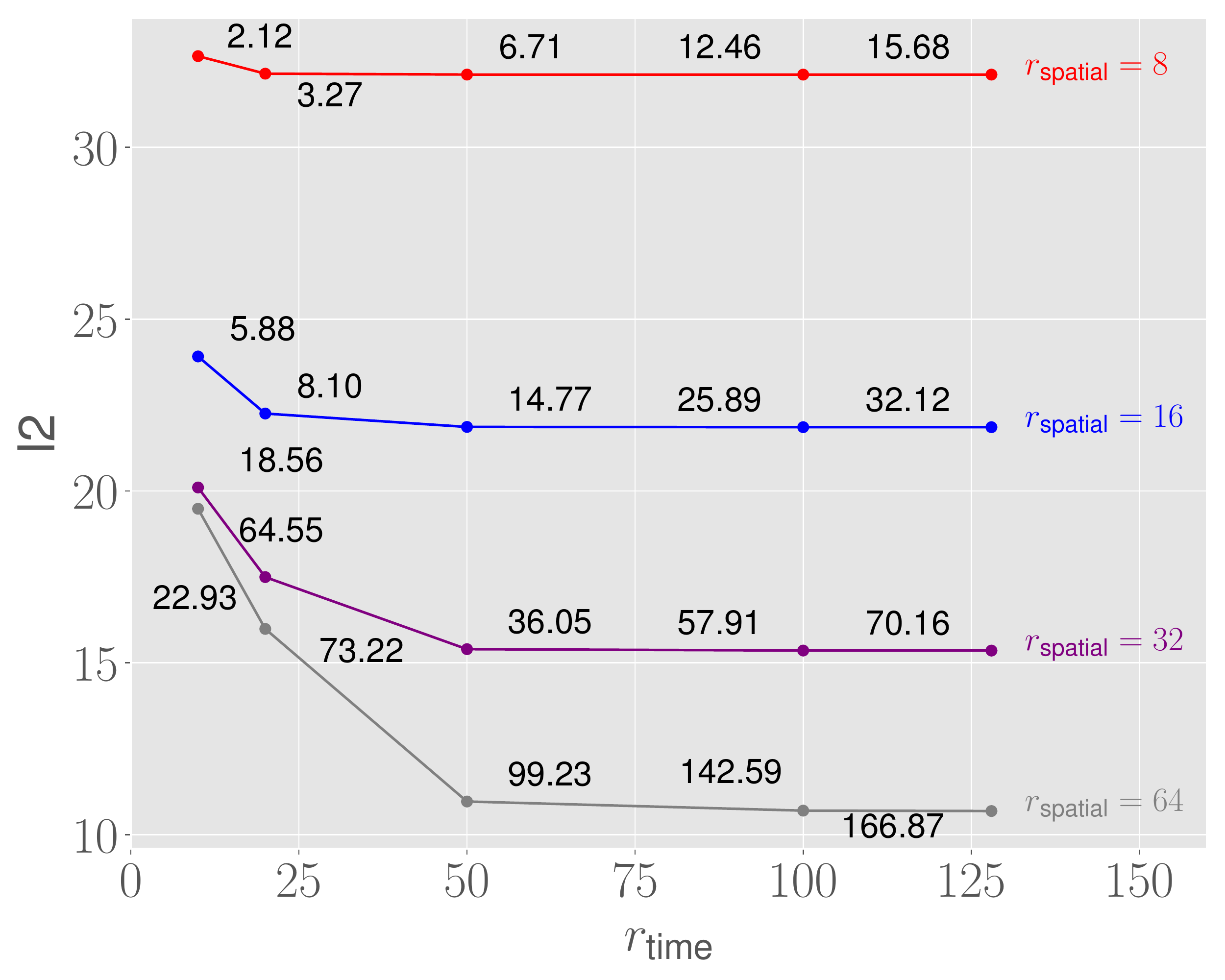}
		\caption{TT}
		\label{fig:tt_l2}
	\end{subfigure}
	\begin{subfigure}[b]{0.49\textwidth}
		\includegraphics[width=\linewidth]{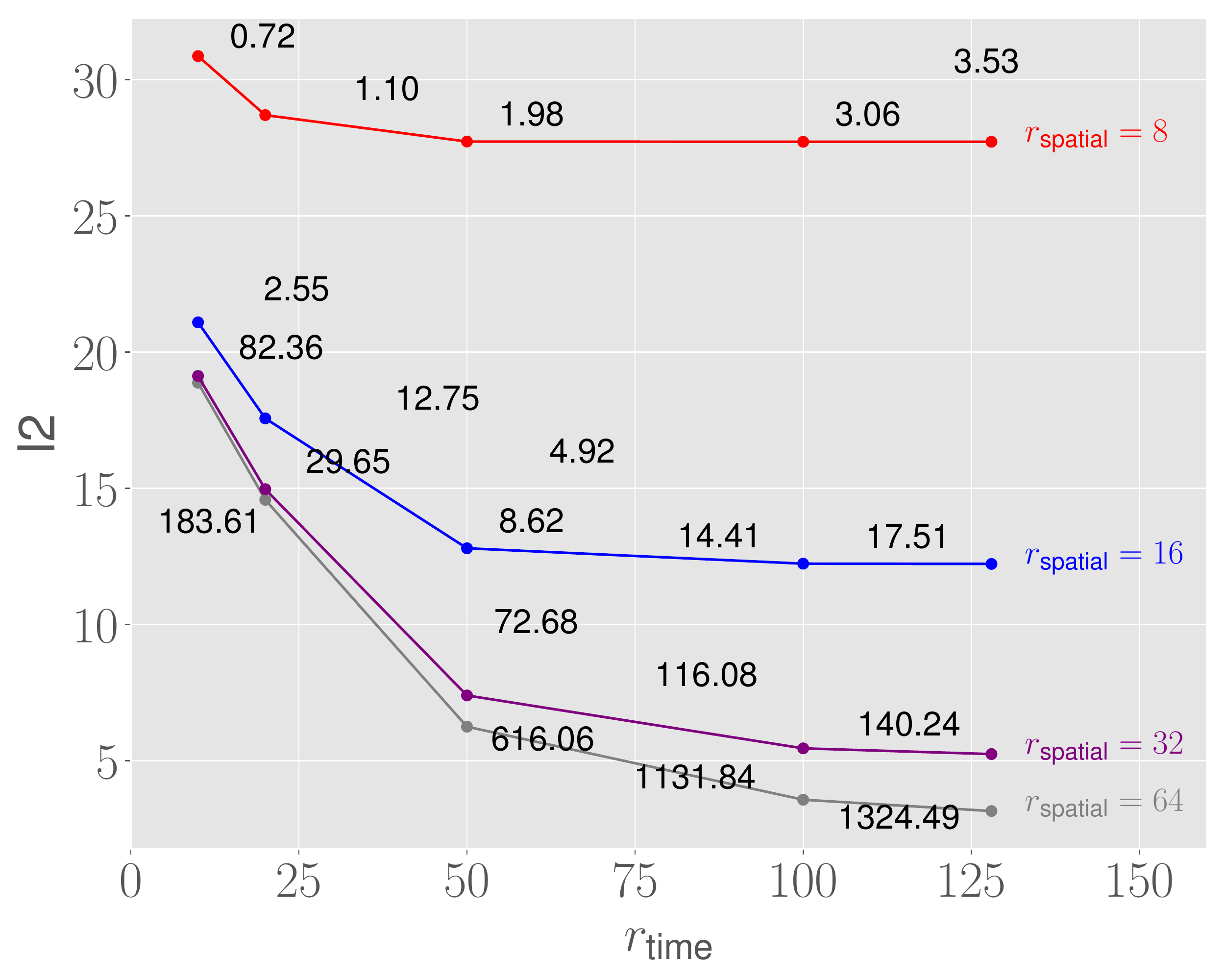}
		\caption{TT Tucker}
		\label{fig:tt_tucker_l2}
	\end{subfigure}
	\caption{Ranks vs. $L_2$ for \textit{longshort-flying-eagle} scene of resolution $512^3$ with 284 time frames. $L_2$ is averaged between frames 1, 142, and 284. Each data point is annotated with the corresponding compression ratio scaled with 1e6.}
	\label{fig:l2}
\end{figure}

\end{document}